\documentclass[journal]{IEEEtran}
\usepackage{amsmath} 
\usepackage{graphicx} 
\usepackage{threeparttable}
\usepackage{multirow}
\usepackage{array}
\usepackage{booktabs}
\usepackage{graphicx}
\usepackage{textcomp}
\usepackage{url}
\usepackage[caption=false,font=footnotesize]{subfig}
\usepackage[table]{xcolor}
\definecolor{outdoorgray}{RGB}{215,215,215} 
\usepackage[ruled,vlined]{algorithm2e}

\title{\LARGE \bf
GR-LIO: A Local Ground-Aware LiDAR–Inertial Odometry System Using Body-to-Ground Height
}
\author{Zhixin Zhang$^{1}$, Yang Song$^{1}$, Liang Zhao$^{2}$, Nathan Shankar$^{3}$, Barry Lennox$^{1}$, Pawel Ladosz$^{3}$
\thanks{This work was partially funded by the Robotics and AI Collaboration (RAICo). (Corresponding author: Pawel Ladosz)}
\thanks{$^{1}$Zhixin Zhang, Yang Song, Barry Lennox are with the Department of Electrical and Electronic Engineering, University of Manchester, Manchester M13 9PL, U.K.
Email: {\tt\small zhixin.zhang@manchester.ac.uk}}
\thanks{$^{2}$Liang Zhao is with the School of Informatics, University of Edinburgh, Edinburgh EH8 9AB, U.K.
Email: {\tt\small liang.zhao@ed.ac.uk}}
\thanks{$^{3}$Nathan Shankar, Pawel Ladosz are with the Department of Mechanical, Aerospace and Civil Engineering, University of Manchester, Manchester M13 9PL, U.K.
Email: {\tt\small pawel.ladosz@manchester.ac.uk}}}

\begin{document}
\maketitle
\thispagestyle{empty}
\pagestyle{empty}

\begin{abstract}
LiDAR–inertial odometry (LIO) is widely used for state estimation in ground-based autonomous mobile robots. However, the geometric constraints provided by the local ground surface remain largely underexploited in existing LIO systems.
This paper proposes a filter-based local ground-aware LIO framework that explicitly incorporates local ground plane geometry into the state estimation process to improve both localization accuracy and computational efficiency. Specifically, a local ground plane is parameterized by the robot orientation and the body-to-ground (B-G) height and continuously propagated within the state estimation process. Based on the proposed B-G geometry model, a propagated local ground plane enables efficient and reliable ground segmentation. The segmented ground points are then incorporated into the filter update through point-to-plane geometric constraints, improving both state estimation accuracy and efficiency. Furthermore, a planar motion update is introduced to exploit the propagated local ground plane as an additional geometric constraint, effectively suppressing vertical drift and improving estimation robustness. To address the initially unknown B-G height, an efficient initialization strategy is developed, followed by an online calibration procedure for continuous refinement. The proposed system is evaluated on several public benchmark datasets and self-collected real-world datasets covering diverse operating scenarios. Experimental results demonstrate that the proposed method consistently outperforms representative LIO methods in terms of both localization accuracy and computational efficiency.
\vspace{0.5em}

\textit{Note to Practitioners}---Many ground robots operate on approximately planar surfaces, yet existing LiDAR-inertial odometry systems often do not explicitly exploit this prior knowledge. The proposed GR-LIO incorporates local ground plane observations and planar motion constraints into the state estimation process, improving both localization accuracy and computational efficiency without requiring additional sensors. Experimental results demonstrate that the proposed method effectively suppresses vertical drift while reducing the computational cost associated with nearest-neighbor searches. The proposed approach can be readily integrated into existing LiDAR-Inertial odometry frameworks and deployed on wheeled robotic platforms operating in both indoor and outdoor environments. The proposed system is designed for ground robots operating under locally planar motion with locally flat ground surfaces, which covers a broad range of indoor and outdoor robotic applications. Future extensions will investigate broader motion models and more challenging terrains while preserving the efficiency of the proposed framework.
\end{abstract}

\section{INTRODUCTION}
LiDAR-based odometry has become a widely adopted solution due to its strong geometric perception capability and robustness in diverse operating environments \cite{c1,c2}. By incorporating inertial measurement unit (IMU) data, LiDAR–Inertial odometry (LIO) systems \cite{c3,c4} further improve motion observability and estimation robustness, especially in structure-less or LiDAR-degenerate scenarios.

\begin{figure}[t]
    \centering
    \subfloat{\includegraphics[width=0.42\textwidth]{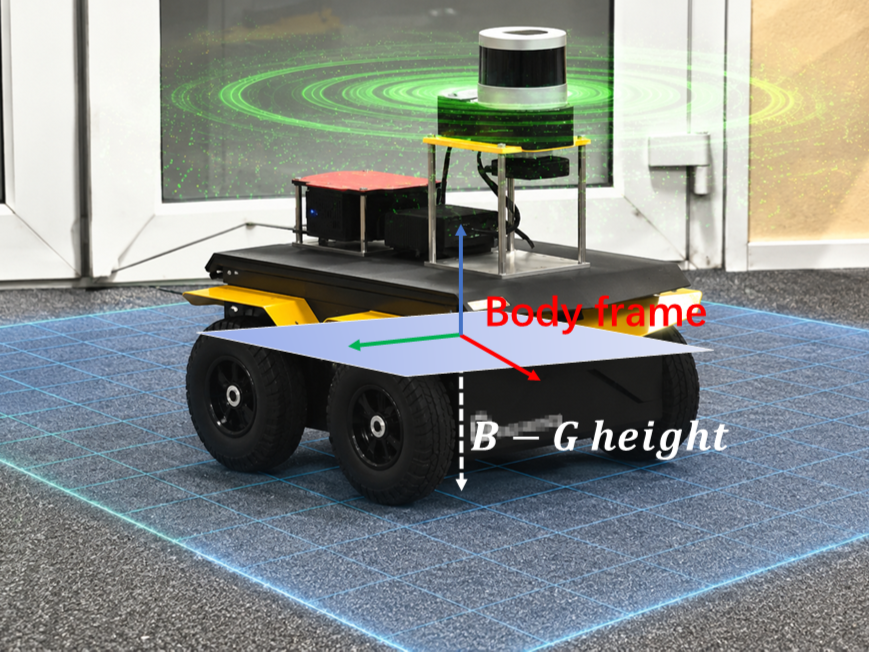}} 
    \caption{Geometric parameterization of the local ground plane using the body-to-ground (B-G) height and robot orientation.}
    \label{fig:first-figure}
\end{figure}

For ground-based mobile robots, the ground surface provides one of the most stable and informative geometric structures in the environment. Accurate ground estimation not only facilitates reliable ground point extraction, but also provides valuable geometric priors for localization and state estimation. Despite the fact that some prior work has considered the ground plane in LiDAR point clouds to improve system performance, such approaches remain limited. For example, LeGO-LOAM \cite{c5} employs a ground segmentation algorithm to improve feature processing efficiency, while GND-LO \cite{c6} models a single plane using multiple ground patches to improve scan-to-scan matching accuracy. However, most existing LiDAR-based frameworks do not explicitly leverage the geometric relationship between the robot body and the ground plane, nor the planar motion characteristics inherent to ground-based platforms. In particular, the planar motion prior and the body–ground geometric coupling remain underutilized, leading to suboptimal estimation consistency and robustness. In this paper, we explicitly model the local ground plane by parameterizing the body-to-ground (B-G) height and the ground normal vector aligned with the robot vertical axis, and incorporate this geometric prior into the LIO state estimation framework to improve performance.

Specifically, a local ground plane representing the terrain in the vicinity of the robot is parameterized by the body-to-ground (B-G) height and the robot orientation. This representation introduces only a single additional parameter, reducing the state dimension and computational complexity compared with explicit ground plane estimation, and remains valid as long as the robot operates on locally planar surfaces, regardless of variations in the ground geometry, as illustrated in Fig.~\ref{fig:first-figure}. Ground points in the current scan are then efficiently segmented and incorporated into the state update through point-to-plane distance observations with respect to the estimated ground plane. Furthermore, a planar motion constraint is introduced based on the local ground plane to suppress drift in the vertical direction. To address an initially unknown or poorly calibrated B-G height, an efficient initialization strategy is proposed, followed by an online calibration procedure for continuous refinement. The main contributions of this work are summarized as follows:
\begin{itemize}
    \item A ground-aware LiDAR-Inertial odometry framework is proposed that explicitly models the local ground plane using the body-to-ground (B-G) height, requiring only one additional state variable.

    \item An efficient local ground point segmentation method is developed, enabling the segmented ground points to be directly incorporated into the state estimation as point-to-plane constraints.

    \item A planar motion update is introduced by exploiting local ground plane, effectively suppressing vertical drift and improving localization accuracy for ground robots.

    \item  A B-G height initialization and online calibration method is proposed to address unknown conditions and compensate for parameter uncertainty.

    \item Extensive experiments on multiple public benchmark datasets and self-collected datasets demonstrate that the proposed method consistently improves localization accuracy and computational efficiency across different environments and LiDAR configurations.
\end{itemize}

The rest of this paper is organized as follows. Section~\ref{sec:relatedwork} reviews the related work. Section~\ref{sec:B-G} presents the proposed B-G geometry model, including ground plane propagation and the planar motion update. Section~\ref{sec:methods} describes the proposed GR-LIO framework, together with the initialization and online calibration of the B-G height. Section~\ref{sec:experiments} presents the experimental evaluation. Finally, Section~\ref{sec:conclusion} concludes the paper.

\section{Related Work}
\label{sec:relatedwork}
In this section, we first review existing LIO methods, which can be broadly categorized into filter-based and optimization-based approaches. We then review previous studies that exploit ground plane priors, including ground segmentation and planar motion constraints, and discuss their limitations in comparison with the proposed method.

\subsection{Optimization-based LIO}
Optimization-based methods have long been a dominant paradigm in LiDAR SLAM due to their robustness and estimation accuracy. In such approaches, IMU measurements are typically incorporated through preintegration factors~\cite{c7}, while LiDAR observations are used to construct geometric constraints for state estimation, such as point-to-edge constraints~\cite{c1} and point-to-plane constraints~\cite{c8}. LIO-SAM~\cite{c3} combines the LOAM-style feature extraction strategy proposed in~\cite{c5} and IMU preintegration for tightly coupled state estimation, where ground and non-ground points are segmented to extract edge and planar features for efficient state estimation. Ye \textit{et al.}~\cite{c9} further proposed a tightly coupled LIO framework based on factor graph optimization that directly fuses raw IMU and LiDAR measurements. LIPS~\cite{c10} represents planes using the closest-point (CP) formulation and incorporates point-to-plane constraints into a factor-graph-based LIO framework. Owing to the modular nature of factor-graph optimization, additional sensing modalities can be readily integrated into the framework. Representative examples include LVI-SAM~\cite{c11}, which incorporates visual measurements, and GR-Fusion~\cite{c12}, which further integrates GNSS and wheel encoder information.

\subsection{Filter-based LIO}

Compared with optimization-based methods, filter-based approaches are generally faster and require lower computational cost, although this efficiency may come at the expense of estimation accuracy. Early filter-based LIO methods, such as LINS~\cite{c13}, adopt a feature extraction pipeline similar to LOAM~\cite{c5} and employ an iterated error-state Kalman filter (IESKF) to reduce linearization errors. Building upon this framework, FAST-LIO~\cite{c14} introduces an equivalent Kalman gain formulation that transforms the measurement dimension into the state dimension, significantly improving computational efficiency. FAST-LIO2~\cite{c4} further eliminates explicit feature extraction by directly constructing point-to-plane constraints from raw LiDAR points. Several subsequent works have focused on improving map management and nearest-neighbor search. For example, Faster-LIO~\cite{c15} replaces the IKD-tree with iVox to further reduce search time while maintaining comparable localization accuracy.

Other LIO systems employ different Kalman filtering formulations. For example, the Multi-State Constraint Kalman Filter (MSCKF) is adopted in \cite{c16,c17,c18} to tightly couple IMU measurements with LiDAR features, whereas IMU, omnidirectional camera, wheel encoder, and LiDAR measurements are integrated within an Extended Kalman Filter (EKF) framework in \cite{c19}. Compared with the conventional EKF and MSCKF, the IESKF performs iterative measurement updates, providing improved handling of nonlinear measurement models and reduced linearization errors. Among the existing filter-based frameworks, FAST-LIO2 provides an efficient and accurate direct LiDAR-Inertial odometry framework. Therefore, the proposed method is developed based on FAST-LIO2 by incorporating local ground plane observations and planar motion constraints into its IESKF framework.

\subsection{Ground Plane in State Estimation}
For ground robots, prior knowledge of the ground plane can often be exploited to improve the performance of LiDAR-based odometry systems. One common application is ground point segmentation, for which several algorithms have been proposed in \cite{c20,c21}. After segmentation, some studies further utilize ground points to accelerate feature processing. For example, the methods presented in \cite{c3,c5} distinguish ground points from non-ground points to improve computational efficiency. In addition, the approach proposed in \cite{c22} treats ground points as stable observations while considering the remaining points as potentially dynamic. GR-LOAM \cite{c23} further exploits the height between the LiDAR sensor and the ground plane for ground segmentation, assuming that the LiDAR sensor is approximately parallel to the ground, an assumption that may not hold in practical deployments. In contrast, the proposed method models the body-to-ground height instead of the LiDAR-to-ground height. Since the B-G height is defined in the robot body frame, it is independent of the LiDAR installation geometry and can be directly estimated as a calibration parameter within the state estimation framework.

Another prior commonly exploited for ground robots is the planar motion assumption, which has been widely adopted in vision-based state estimation systems, like in~\cite{c24,c25,c251,c252}. For LiDAR-based systems, the methods proposed in \cite{c26,c27} incorporate planar surface constraints into LO systems, while the approaches presented in \cite{c28,c29} explicitly exploit ground plane information in LIO systems. However, these methods typically rely on fixed planar motion constraints or require frequent ground plane estimation, which may become unreliable in the presence of uneven terrain, sparse ground observations, or sensor noise. In contrast, the proposed framework explicitly models the local ground plane using the robot orientation and the B-G height. Rather than estimating the ground plane from scratch at every frame, the local ground plane is continuously propagated and refined through the state estimation process. To accommodate uneven or sloped terrain, an efficient local ground-consistency check is introduced to detect changes in the local ground geometry and trigger updates when necessary.

\begin{figure}
    \centering
    \includegraphics[width=0.9\linewidth]{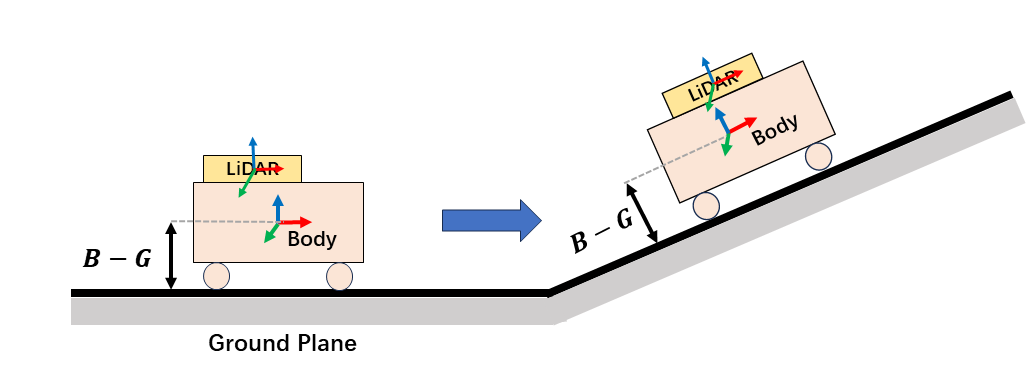}
    \caption{A simple 2-D model of the local ground plane parameterization using the body-to-ground (B-G) height.}
    \label{fig:ground-body}
\end{figure}

\section{Body-Ground Geometry Model}
\label{sec:B-G}

In this section, the Body-Ground (B-G) geometry model used in our system is introduced in detail. In this paper, the IMU frame is defined to be coincident with the robot body frame. Therefore, the IMU frame and the body frame are used interchangeably throughout the paper.

A plane $\mathbf{f}$ in 3D space can be represented by a unit normal vector $\mathbf{n} = {\begin{bmatrix} a & b & c \end{bmatrix}}^{\top}$ and a scalar offset $d$:
\begin{equation}
    \mathbf{f} = \begin{bmatrix} \mathbf{n} \\ d \end{bmatrix}
\end{equation}
For any 3D point $\mathbf{p} = {\begin{bmatrix} x & y & z \end{bmatrix}}^{\top} $ lying on the plane, the following constraint holds:
\begin{equation}
    \mathbf{n}^{\top} \mathbf{p} + d= ax + by + cz + d = 0
    \label{PlaneCons}
\end{equation}

For a mobile robot operating on a ground surface, if the local terrain is approximately planar and the robot undergoes rigid-body motion, the body frame can be assumed to remain locally parallel to the ground plane. Under this assumption, the local ground plane can be parameterized by the Body-Ground height $h$, defined as the signed distance between the robot body center and the ground plane along the body z-axis, as illustrated in Fig.~\ref{fig:ground-body}. The ground plane in the body frame is therefore expressed as:
\begin{equation}
    ^{B}\mathbf{f} = [^{B}\mathbf{n},\, ^{B}d] = \left[ \mathbf{e}_{3}, -h \right]
\end{equation}
where $\mathbf{e}_3 = [0,0,1]^\top$ denotes the unit vector along the vertical direction.

\begin{figure}
    \centering
    \includegraphics[width=0.8\linewidth]{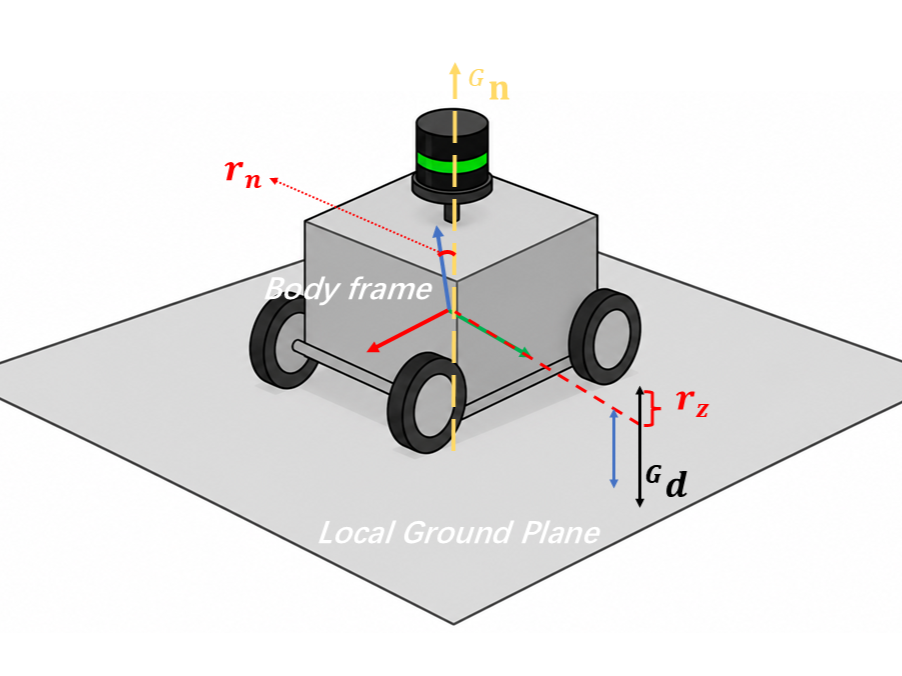}
    \caption{Planar motion residuals used in the proposed state update. $^{G}\mathbf{n}$ and $^{G}d$ denote the normal vector and offset of the ground plane in the global frame, respectively. $r_z$ and $\mathbf{r}_n$ represent the translational and orientation residuals with respect to the local ground plane.}
    \label{fig:planar-residual}
\end{figure}

The distance between the robot body and the ground is adopted for ground modeling instead of the LiDAR–Ground distance. This choice is motivated by the fact that, for ground-based robots, the x–y plane of the body frame is typically aligned with the ground plane, whereas such alignment does not necessarily hold for the LiDAR frame. In practice, LiDAR sensors are often mounted with a tilt angle or multiple LiDAR units are installed to increase sensing coverage. Consequently, the LiDAR-Ground distance adopted in \cite{c12,c23} is dependent on the LiDAR installation geometry, limiting its applicability as a consistent geometric parameter across different sensor configurations.

Once the ground plane $^{B}\mathbf{f} = [^{B}\mathbf{n}, ^{B}d]$ is modeled in the body frame,
it is transformed into the world frame according to the current robot pose as
\begin{equation}
^{G}\mathbf{n} = ^{G}\mathbf{R}_{I_k} \, ^{B}\mathbf{n} = ^{G}\mathbf{R}_{I_k} \, \mathbf{e}_{3}
\label{eq:globalnormal}
\end{equation}
\begin{equation}
^{G}d = ^{G}\mathbf{n}^{\top}\ ^{G}\mathbf{p}_{I_k} + ^{B}d 
= {}^{G}\mathbf{n}^{\top}\ ^{G}\mathbf{p}_{I_k} - h
\label{eq:globaldistance}
\end{equation}
where $^{G}\mathbf{R}_{I_k}$ denotes the rotation matrix from the IMU frame to the world frame at time step $k$, and $^{G}\mathbf{p}_{I_k}$ denotes the position of the IMU frame expressed in the world frame.

This estimated local ground plane is then used for ground point segmentation and state update in the next scan, under the assumption of a locally consistent ground plane. Specifically, for a LiDAR point expressed in the world frame, if its distance to the local ground plane, computed using Eq.~(\ref{PlaneCons}), is below an empirically determined threshold of $0.02~\mathrm{m}$\textbf{}, it is classified as a ground point. 

\begin{figure*}
    \centering
    \includegraphics[width= 0.9\linewidth]{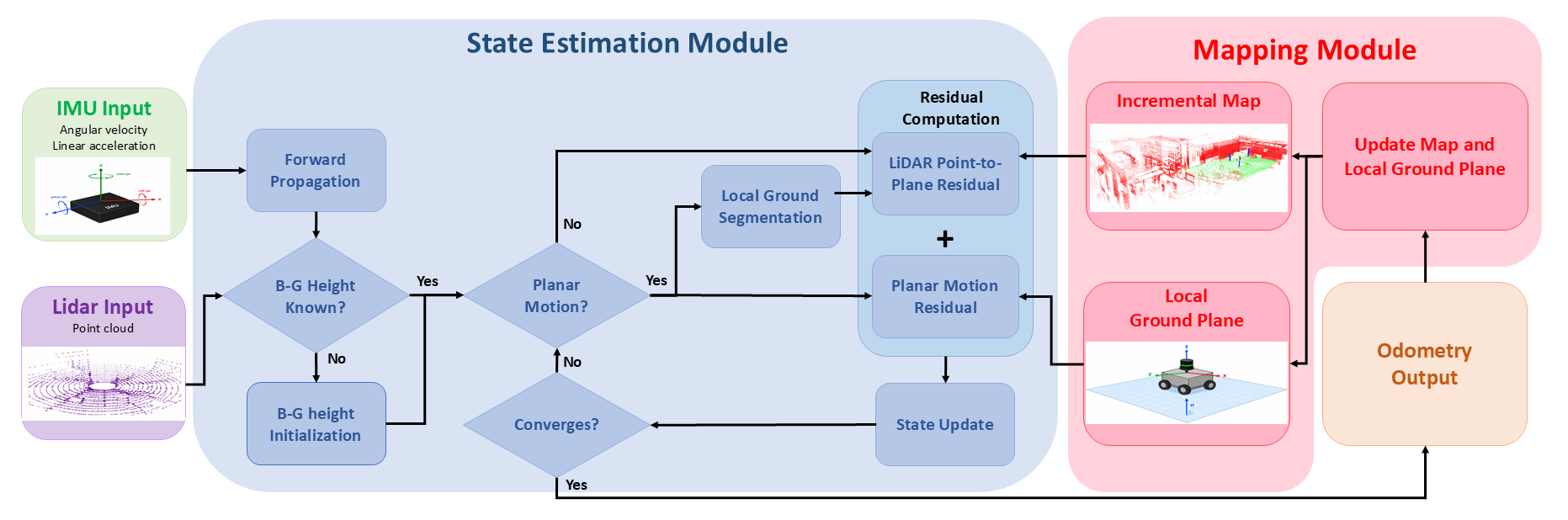}
    \caption{System overview of GR-LIO. The blue block indicates the state estimation process, while the red block represents the mapping module.}
    \label{fig:framework}
\end{figure*}

During the state update step, for each LiDAR point in the current scan that is classified as a ground point and expressed in the world frame, the distance to the local ground plane is directly incorporated as a measurement constraint. For non-ground points, a classical nearest-neighbor search is performed to find the $5$ closest points, which are then used to fit a local plane. The distance between the non-ground point and the fitted plane is used as the measurement residual.

\begin{equation}
r_{p} =
\begin{cases}
^{G}\mathbf{n}_{g}^{\top} \, ^{G}\mathbf{p}_{i} + ^{G}d_{g}, & \text{if } \mathbf{p}_{i} \in \text{ground} \\
^{G}\mathbf{n}_{l}^{\top} \, ^{G}\mathbf{p}_{i} + ^{G}d_{l}, & \text{otherwise}
\end{cases}
\label{eq:gr-feature}
\end{equation}
where $(^{G}\mathbf{n}_g, ^{G}d_g)$ denote the parameters of the global ground plane computed by Eqs.~(\ref{eq:globalnormal}) and (\ref{eq:globaldistance}), while $(^{G}\mathbf{n}_l, ^{G}d_l)$ represent the parameters of the fitted plane for non-ground point neighboring map points.

By leveraging this prior geometric information, ground points can bypass nearest-neighbor search for point-to-plane matching. Instead, they are directly associated with the estimated ground plane and incorporated into the state update. This mechanism also enables effective utilization of distant ground points, where conventional map-based data association often becomes unreliable, thereby increasing the number of valid observations.

In addition, since ground robots typically operate under a planar motion assumption, the robot pose is further constrained by the ground geometry. However, due to IMU noise and insufficient observations, the Z-axis of ground robots often deviates. Specifically, the robot body z-axis expressed in the world frame is enforced to be aligned with the ground normal vector, and the translational displacement along the ground normal direction between the robot body and the ground plane is constrained to remain zero, as shown in Fig.~\ref{fig:planar-residual}.
\begin{equation}
        \mathbf{r}_{n} = ^{G}\mathbf{R}_{B} \mathbf{e}_{3} - ^{G}\mathbf{n}
        \label{r-h}
\end{equation}
\begin{equation}
    r_z = {}^G\mathbf n^\top \left( ^{G}\mathbf{R}_{B} \begin{bmatrix} 0 \\ 0 \\ -h \end{bmatrix}
        + {}^G\mathbf p_B \right) + ^{G}d
\label{r-z}
\end{equation}

\section{SYSTEM IMPLEMENTATION}
\label{sec:methods}

In this section, we present the system overview and implementation details of the proposed GR-LIO framework, including how the proposed B-G geometry model is integrated into the state estimation process, as well as the initialization and online calibration of the B-G height.

\subsection{System Overview}
The proposed framework is illustrated in Fig.~\ref{fig:framework}. The IMU measurements are used for forward propagation at each frame. The propagated state is then used to check whether the planar motion assumption holds. If the assumption is violated, the system falls back to the state update strategy described in \cite{c4}. 
If the planar motion check is satisfied, the LiDAR measurements are segmented into local ground points and non-ground points. For ground points, the residual is computed as the distance to the local ground plane, while for non-ground points, the residual is defined based on the plane fitted from their nearest neighbors. 
In addition, a planar motion constraint is introduced to explicitly constrain the robot to lie on the local ground plane. After the iterative update, both the local ground plane and the map are updated using the odometry output.

\subsection{Filter-based LIO}
\subsubsection{System Dynamic Model}

The state vector of GR-LIO is defined as
\begin{equation}
\mathbf{x}_{k} =
(^{G}\mathbf{R}_{I_k},
\ ^{G}\mathbf{p}_{I_k},
\ ^{G}\mathbf{v}_{I_k},
\ \mathbf{b}_{g_k},
\ \mathbf{b}_{a_k},
\ ^{G}\mathbf{g_k},
\ h)
\end{equation}
where $^{G}\mathbf{v}_{I_k}$ represents the velocity of the robot body expressed in the global frame.
$\mathbf{b}_{g_k}$ and $\mathbf{b}_{a_k}$ denote the gyroscope and accelerometer bias vectors, respectively. 
$^{G}\mathbf{g}$ is the gravity vector expressed in the global frame.
Then the discrete-time state propagation from $t_k$ to $t_{k+1}$ using IMU measurements is given by
\begin{equation}
\mathbf{x}_{k+1} = \mathbf{f}(\mathbf{x}_{k}, \boldsymbol{\omega}_{m_k}, \mathbf{a}_{m_k})
\end{equation}
where $\boldsymbol{\omega}_{m_k}$ and $\mathbf{a}_{m_k}$ denote the gyroscope and accelerometer measurements at time $t_k$. The function $\mathbf{f}(\cdot)$ describes the state propagation model defined as
\begin{equation}
\mathbf{f}(\cdot) =
\mathbf{x}_{k} \oplus
\begin{bmatrix}
\boldsymbol{\omega}_{m_k}-\mathbf{b}_{g_k}-\mathbf{n}_{g_k} \\
{}^{G}\mathbf{v}_{I_k} \\
{}^{G}\mathbf{R}_{I_k}(\mathbf{a}_{m_k}-\mathbf{b}_{a_k}-\mathbf{n}_{a_k})+{}^{G}\mathbf{g_k} \\
\mathbf{n}_{bg_k} \\
\mathbf{n}_{ba_k} \\
\mathbf{0}_{3\times1} \\
0
\end{bmatrix}
\Delta t
\end{equation}
where $\mathbf{n}_{g_k}$ and $\mathbf{n}_{a_k}$ represent zero-mean Gaussian noises, while $\mathbf{n}_{bg_k}$ and $\mathbf{n}_{ba_k}$ denote the white Gaussian noises associated with the gyroscope and accelerometer bias random-walk processes, respectively. The operator $\oplus$ denotes the manifold plus operator, as defined in \cite{c4}. After linearization, the propagation of the error state $\tilde{\mathbf{x}}_{k}$ is given by
\begin{equation}
    \tilde{\mathbf{x}}_{k} \approx 
    \mathbf{F}_{\tilde{\mathbf{x}}}\tilde{\mathbf{x}}_{k-1} + 
    \mathbf{F}_{w}\mathbf{w}_{k-1}
\end{equation}
where $\mathbf{F}_{\tilde{x}}$ and $\mathbf{F}_{w}$ denote the Jacobian matrices of $\mathbf{f}(\cdot)$ with respect to the error state and the process noise, respectively, and $\mathbf{w}_{k-1}$ denotes the process noise associated with the IMU measurements. Denoting the covariance of the current state and the IMU noise as $\mathbf{P}_{I_k}$ and $\mathbf{Q}$, respectively, the covariance propagation can be computed as follows:
\begin{equation}
\hat{\mathbf{P}}_{{k}} =
\mathbf{F}_{\tilde{x}} \mathbf{P}_{k-1} \mathbf{F}_{\tilde{x}}^{\top}
+
\mathbf{F}_{w} \mathbf{Q} \mathbf{F}_{w}^{\top}
\end{equation}

\subsubsection{State Update}
The residuals in Eqs.~(\ref{eq:gr-feature}), (\ref{r-h}), and
(\ref{r-z}) are expected to be zero in the absence of measurement
noise. By stacking these residuals, the nonlinear observation model
at time step $k$ can be written as
\begin{equation}
    \mathbf{0} =
    \mathbf{h}_{k}(\mathbf{x}_{k}) + \mathbf{n}_{o},
\end{equation}
where $\mathbf{h}_{k}(\cdot)$ denotes the stacked observation function and $\mathbf{n}_{o}\sim\mathcal{N}(\mathbf{0},\mathbf{R}_{k})$
represents the observation noise. By linearizing the observation function
around the current nominal state estimate $\hat{\mathbf{x}}_{k}$ as
\begin{equation}
    \mathbf{h}_{k}
    \left(
        \hat{\mathbf{x}}_{k}
        \oplus
        \delta\mathbf{x}_{k}
    \right)
    \approx
    \mathbf{r}_{k}
    +
    \mathbf{H}_{k}
    \delta\mathbf{x}_{k},
\end{equation}
where $\mathbf{H}_{k}$ is its Jacobian matrix with respect to $\tilde{\mathbf{x}}_{k}$ evaluated at zero. The Kalman gain at the $\kappa$-th iteration is computed in information form as~\cite{c14}
\begin{equation}
    \mathbf{K}^{\kappa} =
    \left[ \left(\mathbf{H}_{k}^{\kappa}\right)^{T} \mathbf{R}_{k}^{-1}
        \mathbf{H}_{k}^{\kappa} + \hat{\mathbf{P}}_{k}^{-1}
    \right]^{-1}
    \left(\mathbf{H}_{k}^{\kappa}\right)^{T}
    \mathbf{R}_{k}^{-1},
\end{equation}
This information-form expression is equivalent to the conventional
Kalman gain but avoids the inversion of a matrix whose dimension is
determined by the number of observations. The system state can then be iteratively updated as:
\begin{equation}
\begin{aligned}
    \hat{\mathbf{x}}_{k}^{\kappa+1}
    =
    \hat{\mathbf{x}}_{k}^{\kappa}
    \oplus
        (&-\mathbf{K}^{\kappa}\mathbf{r}_{k}^{\kappa} -
        \left(\mathbf{I} -
            \mathbf{K}^{\kappa}
            \mathbf{H}_{k}^{\kappa}
        \right)
        \left(
            \hat{\mathbf{x}}_{k}^{\kappa}
            \ominus
            \hat{\mathbf{x}}_{k}
        \right))
\end{aligned}
\end{equation}
Subsequently, each LiDAR point $^{L}\mathbf{p}_{j}$ in the current scan is transformed into the global frame using the updated state as
\begin{equation}
^{G}\mathbf{p}_{j} = {}^{G}\mathbf{T}_{I_k}\,{}^{I}\mathbf{T}_{L}\,{}^{L}\mathbf{p}_{j}.
\end{equation}
The local ground plane is then transformed into the global frame using the updated state and the estimated B-G height, where the global ground plane normal vector and offset are computed according to (\ref{eq:globalnormal}) and (\ref{eq:globaldistance}), respectively. The transformed LiDAR points and local ground plane are subsequently used for the subsequent update.

\subsection{B-G Height Initialization and Online Calibration}
The proposed framework relies on an accurate B-G height, which may not always be available in practice due to unknown or inaccurate initial calibration. To address this issue, we propose a B-G height initialization method to estimate the parameter from an unknown initial condition, followed by an online calibration strategy that continuously refines the estimate during operation.

\subsubsection{Initialization}
As discussed in Section~\ref{sec:B-G}, the ground plane normal vector $\mathbf{n}$ can be directly determined from the robot orientation and therefore does not require explicit initialization. Consequently, the B-G height $h$ is the only parameter that needs to be estimated during the initialization stage. The proposed initialization procedure is summarized in Alg.~\ref{alg:bg_init}.

\begin{algorithm}[!t]
\caption{B-G Height Initialization}
\KwIn{Point cloud $\mathbf{P}$, robot rotation $\mathbf{R}$}
\KwOut{Estimated B-G height $h$}

Compute ground normal vector $\mathbf{n}=\mathbf{R}[0,0,1]^\top$\;
Normalize $\mathbf{n}$\;

Initialize height set $\mathcal{H}\leftarrow\emptyset$\;

\textbf{// Select the lowest points}\;
\For{each point $\mathbf{p}_i$ in $\mathbf{P}$}{
    \If{$\|\mathbf{p}_i\|>r_{\max}$}{
        continue\;
    }
    $h_i \leftarrow \mathbf{n}^{\top}\mathbf{p}_i$\;
    \If{$h_i>0$}{
        continue\;
    }
    Store $h_i$ in $\mathcal{H}$\;
}

Sort $\mathcal{H}$ in ascending order\;
$N_{\mathrm{LPR}}\leftarrow \min(50,|\mathcal{H}|)$\;
Select the lowest $N_{\mathrm{LPR}}$ values from $\mathcal{H}$ to form $\mathcal{H}_{\mathrm{LPR}}$\;

\textbf{// RANSAC-based B-G height estimation}\;
Initialize best inlier count $C^*\leftarrow -1$\;

\For{$k=1$ to $K$}{
    Randomly sample $h_0$ from $\mathcal{H}_{\mathrm{LPR}}$\;
    $\mathcal{I}\leftarrow\{i\mid |h_i-h_0|<\tau,\ h_i\in\mathcal{H}_{\mathrm{LPR}}\}$\;

    \If{$|\mathcal{I}|>C^*$}{
        $C^*\leftarrow|\mathcal{I}|$\;
        $\mathcal{I}^*\leftarrow\mathcal{I}$\;
    }
}

$h_{\mathrm{LPR}}\leftarrow \frac{1}{|\mathcal{I}^*|}\sum_{i\in\mathcal{I}^*}h_i$\;

$h\leftarrow h_{\mathrm{LPR}} + h_{\mathrm{LB}}$\;

\Return $h$\;

\label{alg:bg_init}
\end{algorithm}

First, points within a local neighborhood satisfying $|\mathbf{p}_i|<r_{\max}$ are considered, where $r_{\max}=4~\mathrm{m}$ in our implementation. Following the strategy in~\cite{c21}, these points are projected onto the ground normal direction and sorted according to their projected heights. The lowest $N_{\mathrm{LPR}}$ points are then selected as candidate ground points. Unlike~\cite{c21}, which estimates both the ground plane normal and its offset, the proposed method estimates only the plane offset, since the ground normal is directly determined by the robot orientation. Consequently, the initialization reduces to estimating a single parameter corresponding to the B-G height.

The selected candidate points are subsequently used to estimate the B-G height within a Random Sample Consensus (RANSAC)~\cite{c291} framework to reject outliers. Since the ground normal is known, RANSAC is performed only on the one-dimensional projected heights. The optimal B-G height is obtained by minimizing the residuals over the inlier set:
\begin{equation}
h^*=\arg\min_h\sum_{i\in\mathcal{I}}\left(\mathbf{n}^{\top}\mathbf{p}_i-h\right)^2.
\end{equation}
where $\mathbf{p}_i$ denotes the $i$-th candidate ground point and $\mathcal{I}$ is the inlier set determined by RANSAC. In our implementation, the number of lowest points used for initialization is set to $N_{\mathrm{LPR}}=50$, the maximum number of RANSAC iterations is set to $K=20$, and the inlier distance threshold is set to $\tau=0.02~\mathrm{m}$. If the number of inliers is smaller than $N_{\min}$, the average height of the candidate ground points is used as a fallback estimate. Finally, a constant vertical offset between the LiDAR and the IMU frames, $h_{\mathrm{LB}}$ is added to convert the estimated LiDAR-to-ground height into the B-G height. This initialization requires only a single scan and assumes that the robot is initially located on a locally planar ground surface.
\subsubsection{Online Calibration}
Although the B-G height can be initialized as described above, the estimate may still be inaccurate. Therefore, an online calibration procedure is introduced to further refine the estimate.

As shown in Eq.~(\ref{eq:gr-feature}), the ground point observation is governed by the local ground plane. By incorporating Eqs.~(\ref{eq:globalnormal}) and (\ref{eq:globaldistance}), the observation function can be expressed in terms of the B-G height $h$:
\begin{equation}
    \mathbf{r}_{p} = (^{G}\mathbf{R}_{I_{k-1}} {}\mathbf{e}_{3})^{\top} {}^{G}\mathbf{p}_{i} + 
    (^{G}\mathbf{R}_{I_{k-1}} {}\mathbf{e}_{3})^{\top}{}^{G}\mathbf{p}_{I_{k-1}} - h
\end{equation}
where $^{G}\mathbf{R}_{I_{k-1}}$ represents the robot orientation at time $k-1$ expressed in the world frame. The pose at time $k-1$ is used instead of the current state at time $k$ because the local ground plane is estimated using the previously updated state and the B-G height. Therefore, the local plane depends only on the B-G height $h$ and is independent of the current pose variables. ${}^{G}\mathbf{p}_{i}$ represents the $i$-th LiDAR point in the current scan expressed in the global frame, which is computed as:
\begin{equation}
{}^{G}\mathbf{p}_{i}
=
{}^{G}\mathbf{R}_{I_k}
\left(
{}^{I}\mathbf{R}_{L}
{}^{L}\mathbf{p}_{i}
+
{}^{I}\mathbf{t}_{L}
\right)
+
{}^{G}\mathbf{t}_{I_k}.
\end{equation}

Then the Jacobian of the residual with respect to the B-G height $h$ can be directly obtained as
\begin{equation}
\frac{\partial r_{p}}{\partial h} = -1.
\end{equation}
Since the proposed observation model only augments the local ground plane with the B-G height parameter, the Jacobians with respect to the remaining state variables are identical to those in~\cite{c4}.

\subsection{Dealing with Complex Road Conditions}
\begin{figure}
    \centering
    \subfloat[Sloped terrain]{\includegraphics[width=0.9\linewidth]{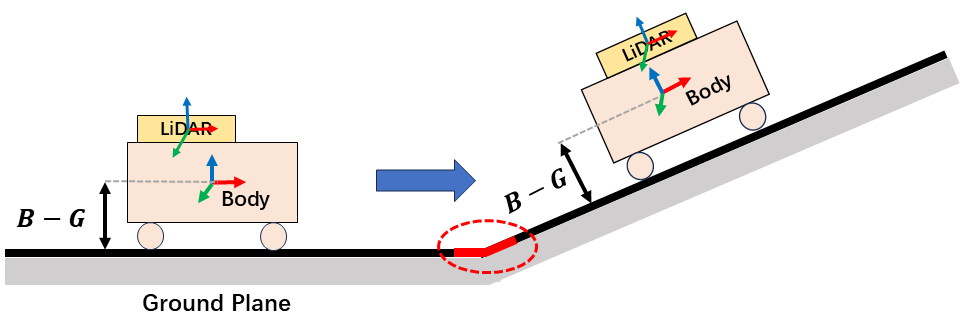}}
    \\ 
    \subfloat[Uneven terrain]{\includegraphics[width=0.9\linewidth]{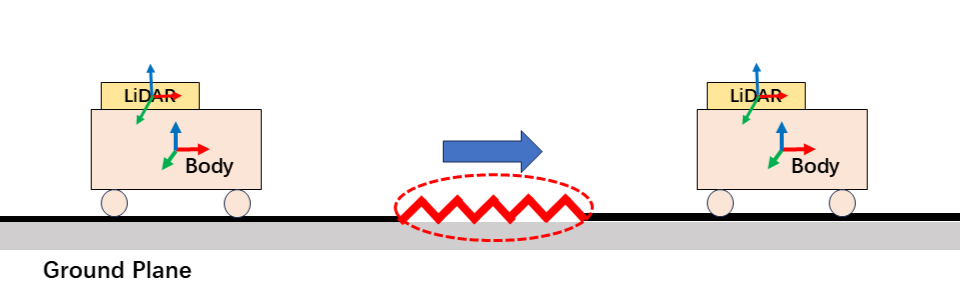}}
    \caption{Examples of situations where the planar motion assumption is violated. The red highlighted regions indicate areas where the local ground geometry changes and the B-G based local ground model becomes invalid.}
    \label{fig:ComplexCondition}
\end{figure}

In real-world applications, the planar motion assumption may not always hold, particularly during long-term outdoor operation. Directly applying the B-G-based parameterization under such conditions can degrade localization performance. For example, when the robot transitions between different ground surfaces, such as moving onto a slope, the local ground geometry changes and temporarily violates the planar motion assumption, as shown in Fig.~\ref{fig:ComplexCondition}(a). Similarly, in uneven environments, local ground characteristics may vary frequently, leading to inconsistencies between the propagated ground plane and the actual terrain, as illustrated in Fig.~\ref{fig:ComplexCondition}(b).

To mitigate this issue, we propose a planar motion consistency check to determine whether the B-G model remains valid during operation. Specifically, after IMU forward propagation in the state estimation process, a metric is computed to evaluate whether the planar motion assumption holds. The metric is defined as the deviation of the robot position from the estimated ground plane along the ground normal direction:
\begin{equation}
e_{\text{drift}} = {}^{G}\mathbf{n}^{\top} \, {}^{G}\mathbf{p}_{B} - {}^{G}d
\end{equation}

If the metric exceeds a predefined threshold (0.01~m in our implementation), the planar motion assumption is considered invalid. In this case, the proposed ground-aware update is disabled, and the system falls back to the standard state update strategy~\cite{c4}.

\begin{figure}
    \centering
    \subfloat{\includegraphics[width=0.9\linewidth]{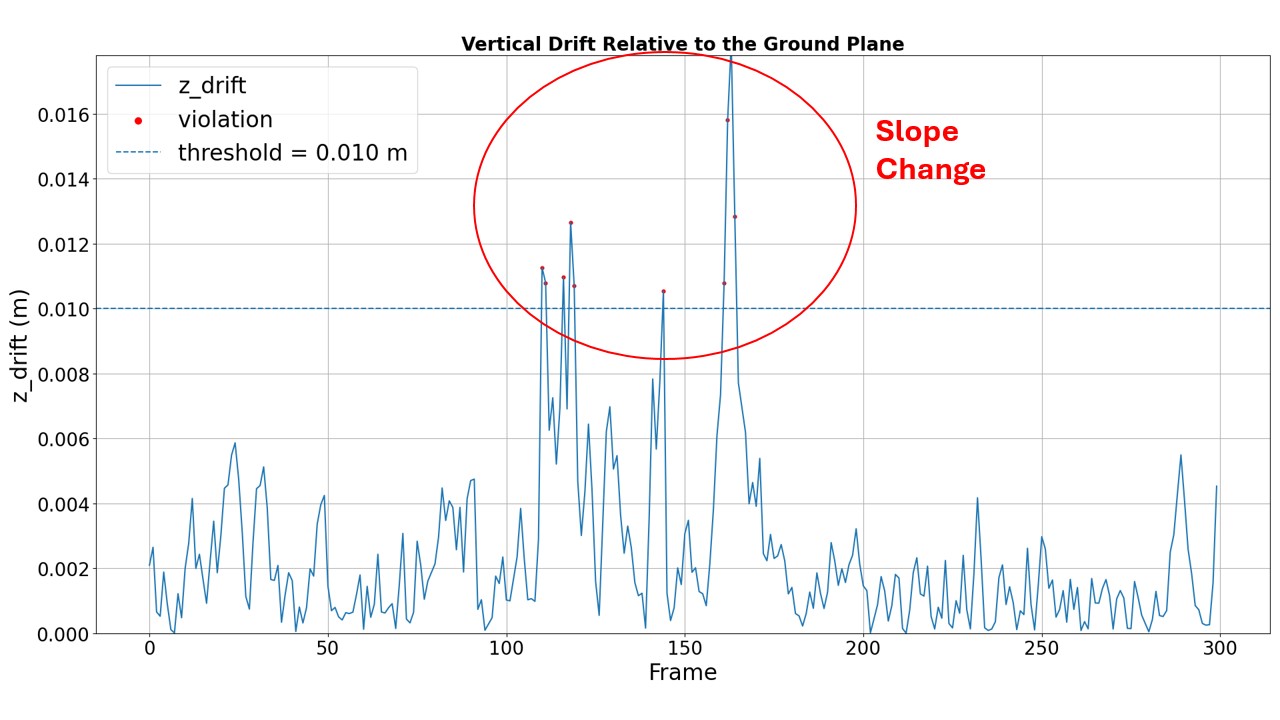}}
    \\ 
    \subfloat{\includegraphics[width=0.9\linewidth]{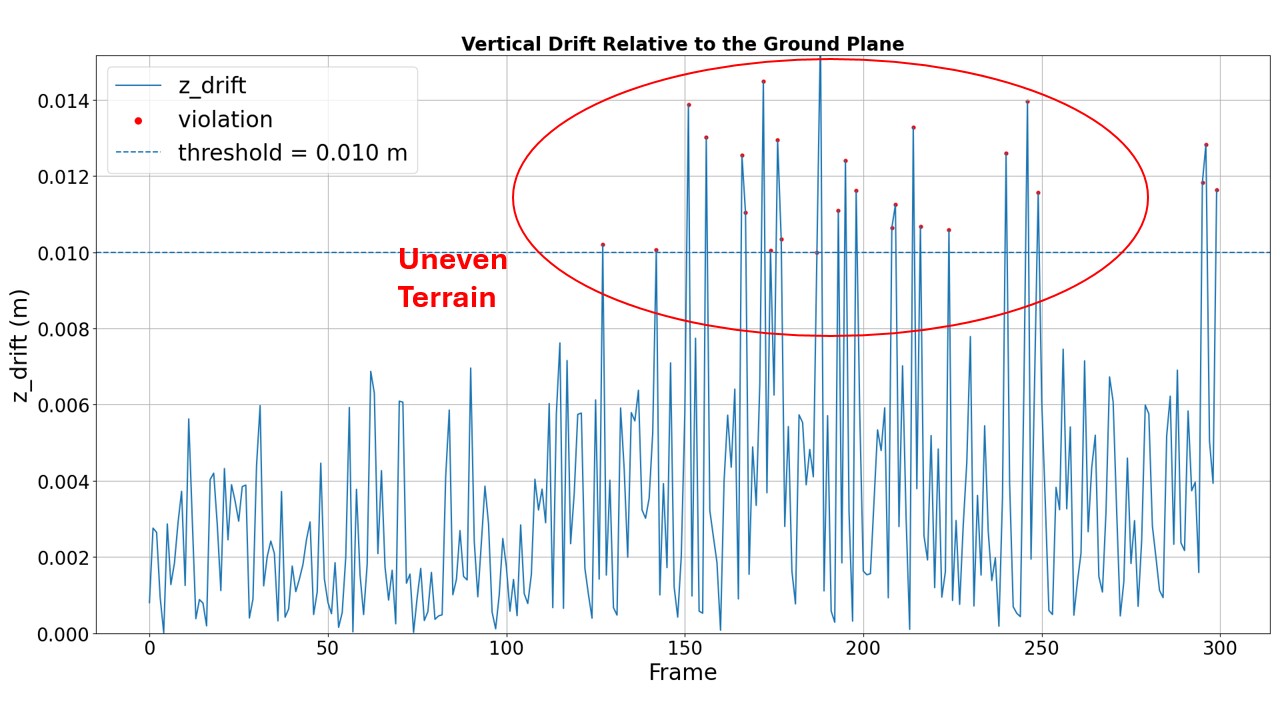}}
    \caption{Vertical drift relative to the estimated ground plane during slope transitions and uneven-terrain traversal. The dashed line indicates the threshold used for planar motion checking, while red markers denote samples that exceed the threshold.}
    \label{fig:Z_drift}
\end{figure}

Fig.~\ref{fig:Z_drift} shows two representative scenarios in which the vertical drift changes during operation. Fig.~\ref{fig:Z_drift}(a) illustrates a transition to a sloped terrain, while Fig.~\ref{fig:Z_drift}(b) corresponds to an uneven-terrain scenario. It can be clearly observed that the vertical drift increases significantly whenever the local ground plane assumption is violated. In both scenarios, the drift quickly exceeds the predefined threshold, providing a reliable indicator for detecting invalid ground-plane assumptions and triggering the fallback update strategy.

\section{EXPERIMENTS}
\label{sec:experiments}
\subsection{Experiment Setup}
\subsubsection{Public Datasets}
We evaluated the proposed system on two public datasets, M2DGR~\cite{c30} and M3DGR~\cite{c31}, both collected using ground robot platforms equipped with LiDAR and IMU sensors under diverse enviroments. These datasets cover a wide range of environments, including both indoor and outdoor scenes, providing suitable benchmarks for evaluating the robustness and generalization capability of the proposed method. M2DGR was collected using a spinning LiDAR sensor (Velodyne, 360$^\circ$ horizontal field of view) together with an external IMU (Handsfree A9, 9-axis, 150~Hz). In contrast, M3DGR employs solid-state LiDAR sensors, including the Livox Avia (70.4$^\circ$ field of view), together with its built-in 6-axis IMU operating at 200~Hz. For both datasets, the B-G height was provided as ground-truth information.
\subsubsection{Self-Collected Dataset}
\begin{figure}[t]
    \centering
    \subfloat[Experimental platform.]{\includegraphics[width=0.495\linewidth]{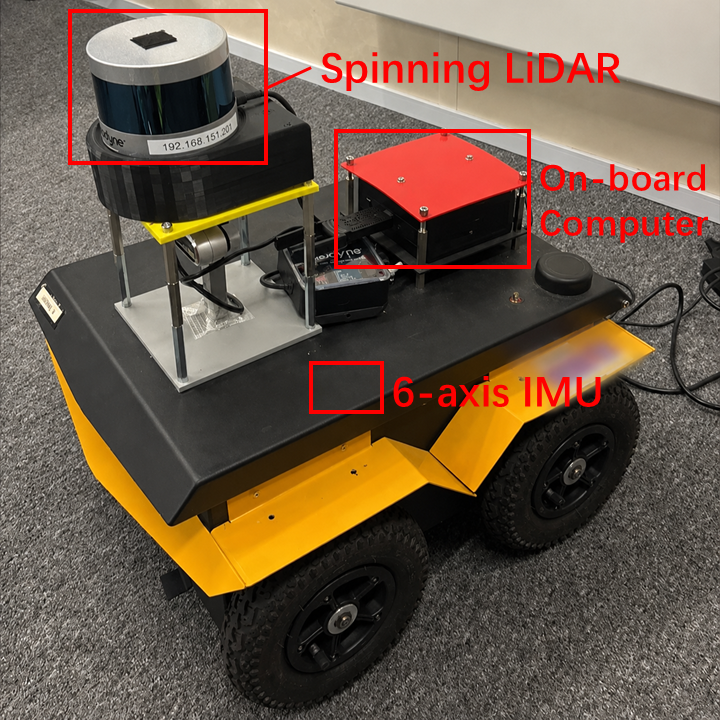}}
    \hfill
    \subfloat[Robot during data collection.]{\includegraphics[width=0.495\linewidth]{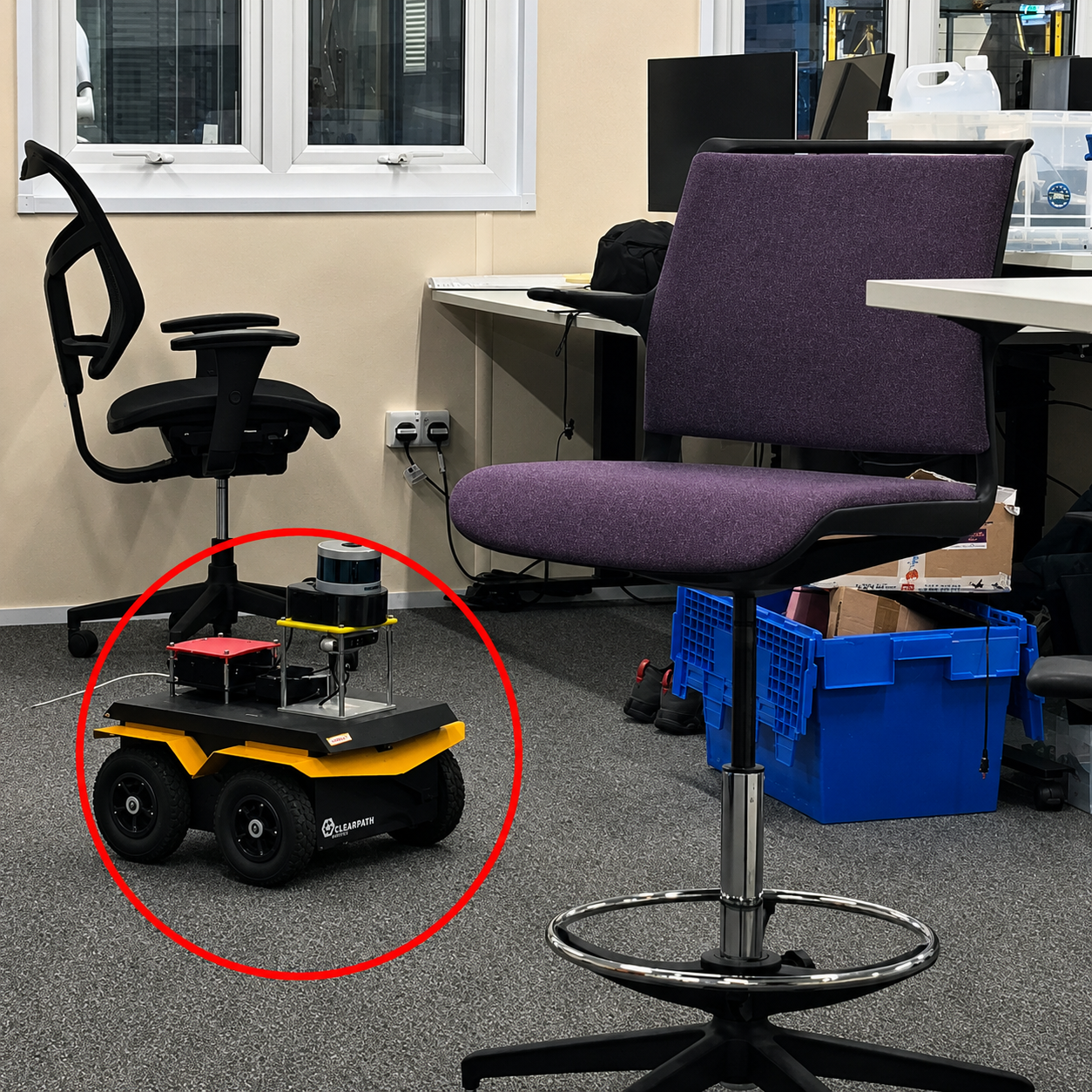}}   
    \\
    \subfloat[Floor plan of the test environment, where the black arrows indicate the start and end points and direction of each sequence.]{
        \includegraphics[width=0.85\linewidth]{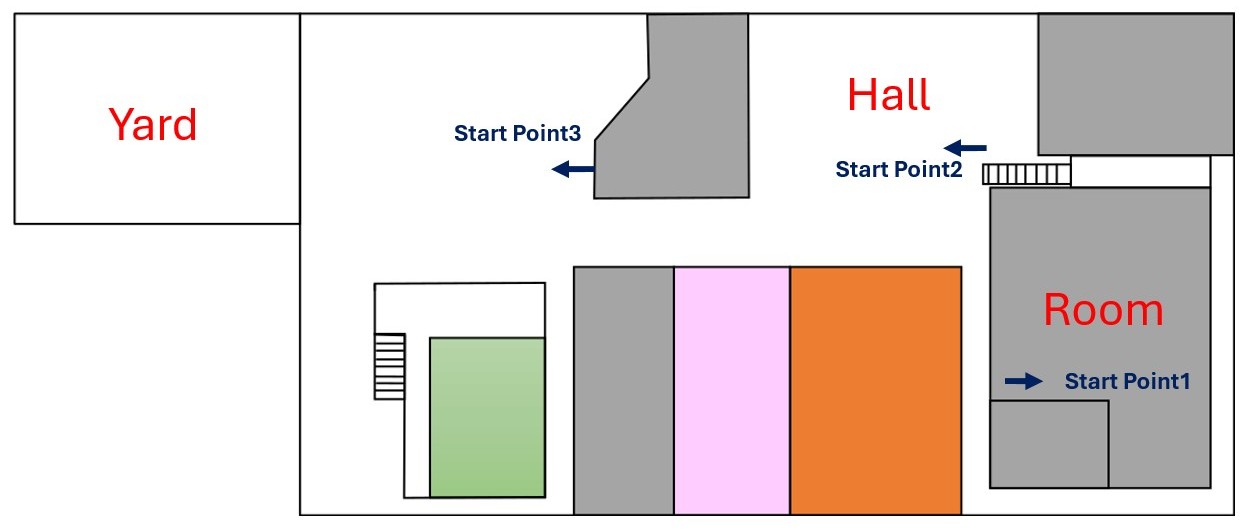}}
    \caption{Experimental platform and data collection environments.}
    \label{fig:exp-platform}
\end{figure}

To further evaluate the proposed system under practical operating conditions, we collected a self-built dataset using the experimental platform shown in Fig.~\ref{fig:exp-platform}(a). The platform consists of a spinning LiDAR sensor (Velodyne VLP-16 Puck) and a 6-axis IMU mounted on a wheeled robot, with all sensor parameters calibrated prior to data collection.
The dataset contains three representative sequences collected in different environments, including indoor corridors, large halls, and indoor-to-outdoor transitions. The corresponding floor plan and representative scenes are shown in Fig.~\ref{fig:exp-platform}(c). In addition, the B-G height was manually measured and used as ground truth for evaluating the proposed initialization and online calibration methods.

\subsubsection{Benchmark and Metrics}
We compare the proposed system with several state-of-the-art LiDAR--inertial odometry methods, including LINS~\cite{c13}, LIO-SAM~\cite{c3}, and FAST-LIO2~\cite{c4}. LINS and FAST-LIO2 are based on the IESKF, whereas LIO-SAM adopts a graph-based optimization framework.
In addition, a variant denoted as \textbf{GR-LIO (FP)} is introduced, in which the local ground plane is estimated during initialization and subsequently kept fixed throughout the trajectory, instead of being represented and continuously adapted using the proposed B--G parameterization. This variant is included to evaluate the benefit of the proposed ground plane parameterization in accommodating changes in the local ground geometry. A second variant without the planar motion update (\textbf{w/o PU}) is included to evaluate the contribution of the planar motion constraint described in Section~\ref{sec:B-G}. Since the publicly available implementations of LINS and LIO-SAM only support spinning LiDAR sensors, comparisons on M3DGR are limited to FAST-LIO2. In contrast, M2DGR and the self-collected datasets employ spinning LiDAR sensors, allowing all three baseline methods to be included.

For the M2DGR and M3DGR datasets, localization accuracy is evaluated by comparing the estimated trajectories with the corresponding ground-truth trajectories using the evo toolbox~\cite{c32}. The root mean square error (RMSE) of the absolute trajectory error (ATE) is adopted as the primary localization accuracy metric, where a lower value indicates higher localization accuracy.

\subsubsection{Implementation}
Both the proposed system and the baseline methods are implemented in C++. Unless otherwise specified, all methods use the default noise parameters provided by the corresponding datasets. To ensure a fair comparison, loop closure and all online calibration modules are disabled for all evaluated methods. Experiments are conducted on a workstation equipped with an Intel Core i7 processor, 32~GB RAM, and an NVIDIA RTX A1000 GPU with 6~GB memory, running Ubuntu 20.04.

\subsection{Ground Point Segmentation}
\begin{figure*}
    \subfloat[hall 01]{\includegraphics[width= 1.0\linewidth]{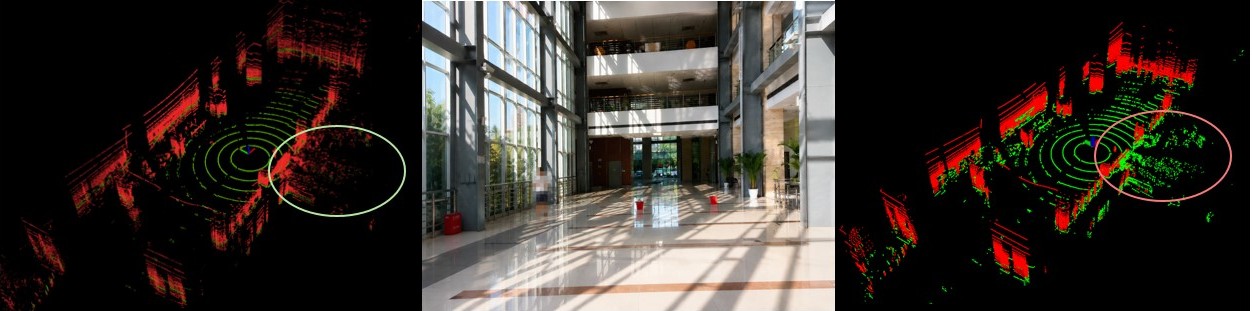}}
    \\
    \subfloat[door 01]{\includegraphics[width= 1.0\linewidth]{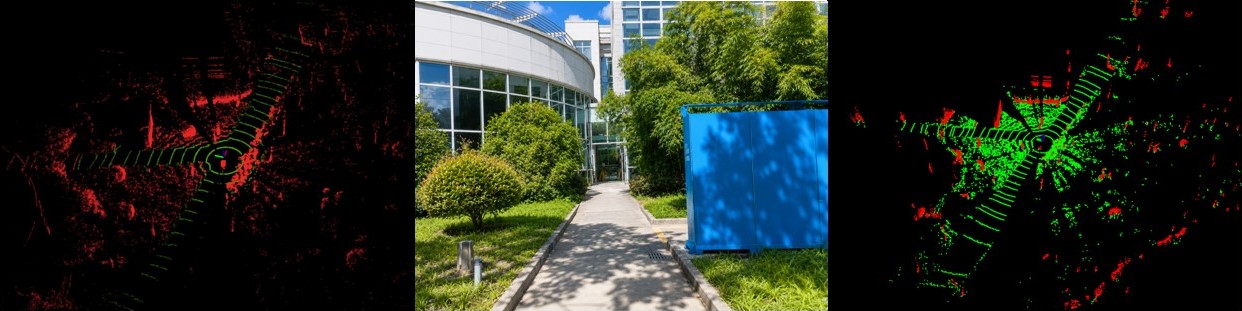}}
    \\
    \subfloat[gate 01]{\includegraphics[width=1.0\linewidth]{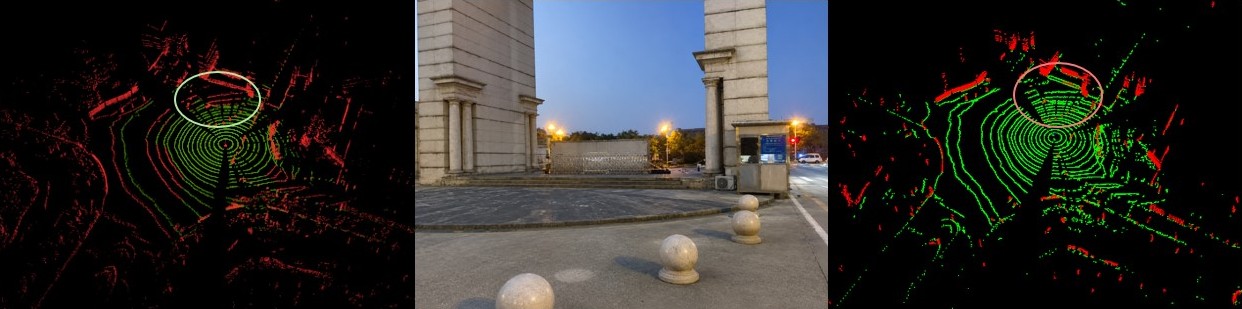}}
    \\
    \subfloat[street 03]{\includegraphics[width=1.0\linewidth]{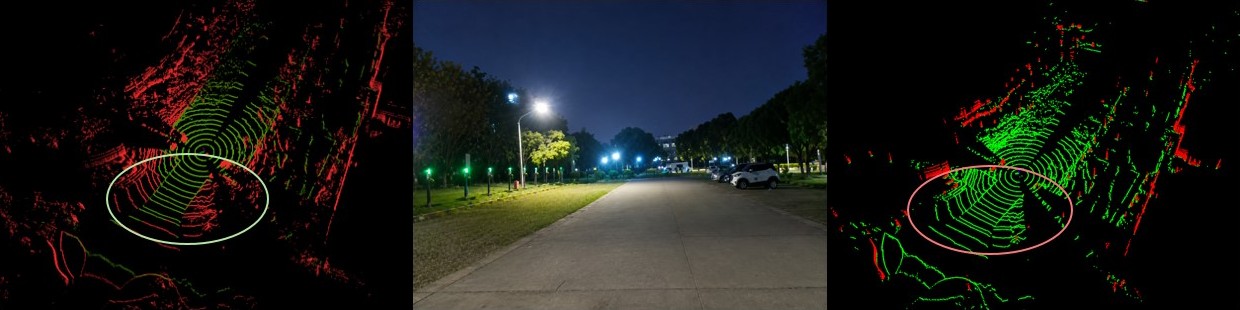}}
    \caption{Qualitative comparison of ground segmentation results between GR-LIO and LeGO-LOAM on representative sequences from the M2DGR dataset. For each sequence, the left column shows the segmentation results of GR-LIO, the center column presents an image of the corresponding real-world environment for reference, and the right column shows the segmentation results of LeGO-LOAM. The major differences between the two methods are highlighted with green and red circles to facilitate visual comparison. It can be observed that LeGO-LOAM produces more wromng ground segmentations than GR-LIO.}
    \label{fig:GroundSeg}
\end{figure*}

To evaluate the effectiveness of the proposed B--G-based ground segmentation method, we compare it with the ground segmentation algorithm adopted in LeGO-LOAM~\cite{c5}. Representative segmentation results on different M2DGR sequences are shown in Fig.~\ref{fig:GroundSeg}. Since the segmented ground points are directly incorporated into the state update as point-to-plane observations, more accurate ground segmentation provides more reliable geometric constraints for state estimation and is expected to improve localization performance.

As shown in Fig.~\ref{fig:GroundSeg}, GR-LIO consistently produces cleaner and more complete ground segmentation results across different environments. Compared with LeGO-LOAM, fewer non-ground structures are incorrectly classified as ground, while more valid ground points are preserved. This improvement is particularly evident in scenarios with local terrain variations, such as gentle slopes, grassy areas, and curbs, where LeGO-LOAM tends to incorrectly classify these regions as ground. In contrast, the proposed method leverages the propagated local ground plane derived from the B-G model, allowing it to better distinguish true ground points from local terrain variations and preserve more reliable ground observations.

\begin{table}[t]
\centering
\caption{Average execution time of the ground segmentation module on representative M2DGR sequences (ms).}
\label{tab:ground_seg_time}
\setlength{\tabcolsep}{10pt}
\renewcommand{\arraystretch}{1.2}
\begin{tabular}{c|cc}
\toprule
\textbf{Sequence} & \textbf{GR-LIO (Ours)} & \textbf{LeGO-LOAM~\cite{c5}} \\
\midrule
hall 01 & 0.45 & 7.55 \\
door 01 & 0.48 & 9.05 \\
gate 01 & 0.43 & 7.89 \\
street 03 & 0.44 & 7.98 \\
\bottomrule
\end{tabular}
\end{table}

To further evaluate the computational efficiency of the proposed ground segmentation method, we compare its average processing time per LiDAR frame with that of LeGO-LOAM. As summarized in Table~\ref{tab:ground_seg_time}, GR-LIO requires only 0.45 ms per frame on average, whereas LeGO-LOAM requires approximately 8 ms, corresponding to an approximately 18-fold reduction in processing time. This efficiency gain is primarily attributed to the propagated local ground plane, which enables direct classification of ground points using point-to-plane distances. In contrast, LeGO-LOAM performs scan-dependent geometric comparisons between neighboring LiDAR measurements and projects the point cloud onto a range image, resulting in substantially higher computational overhead. Furthermore, LeGO-LOAM relies on LiDAR-specific parameter tuning to accommodate different sensor configurations, whereas the proposed method performs ground segmentation without any sensor-dependent parameter adjustment. These results demonstrate that the proposed method achieves both efficient and reliable ground segmentation while offering improved portability across different LiDAR platforms.

\subsection{LIO with Known G-B Height}
In this section, we evaluate the localization accuracy of the proposed LIO system under the condition that the B-G height is known and accurate. This assumption holds in most practical scenarios, as it can be readily measured. 

Compared with the relatively stable and approximately planar motion in indoor environments, outdoor scenarios are typically more challenging due to uneven terrain and varying slopes. For clarity, the outdoor sequences are indicated by a gray background in the tables. The best and second-best results are highlighted in \textbf{bold} and \underline{underlined}, respectively.

\subsubsection{M2DGR Dataset}
\begin{table*}[t]
  \centering
  \caption{RMSE comparison of different algorithms and GR-LIO variants on the M2DGR dataset
  (APE$_t$: m, RPE$_t$: m).}
  \label{tab:m2dgr_accuracy}
  \setlength{\tabcolsep}{5pt}
  \renewcommand{\arraystretch}{1.15}

  \begin{threeparttable}
  \resizebox{\textwidth}{!}{%
  \begin{tabular}{c|cc|cc|cc|cc|cc|cc}
    \toprule
    \multirow{2}{*}{\textbf{Sequence}} &
    \multicolumn{2}{c|}{\textbf{LINS~\cite{c13}}} &
    \multicolumn{2}{c|}{\textbf{LIO-SAM~\cite{c3}}} &
    \multicolumn{2}{c|}{\textbf{FAST-LIO2~\cite{c4}}} &
    \multicolumn{2}{c|}{\textbf{GR-LIO (FP)}} &
    \multicolumn{2}{c|}{\textbf{GR-LIO(w/o PU)}} &
    \multicolumn{2}{c}{\textbf{GR-LIO}} \\
    
    \cmidrule(lr){2-3}
    \cmidrule(lr){4-5}
    \cmidrule(lr){6-7}
    \cmidrule(lr){8-9}
    \cmidrule(lr){10-11}
    \cmidrule(lr){12-13}
    
    & APE$_t$ & RPE$_t$
    & APE$_t$ & RPE$_t$
    & APE$_t$ & RPE$_t$
    & APE$_t$ & RPE$_t$
    & APE$_t$ & RPE$_t$
    & APE$_t$ & RPE$_t$ \\
    \midrule

    \rowcolor{outdoorgray}
    gate 01 &
    0.296 & 0.353 &
    \textbf{0.164} & 0.298 &
    0.179 & \underline{0.145} &
    0.173 & 0.147 &
    \underline{0.172} & \textbf{0.144} &
    \underline{0.172} & \textbf{0.144} \\

    \rowcolor{outdoorgray}
    gate 02 &
    \multicolumn{2}{c|}{FAIL} &
    0.438 & 0.288 &
    \underline{0.325} & \underline{0.142} &
    0.329 & 0.096 &
    0.328 & \underline{0.142} &
    \textbf{0.320} & \textbf{0.140} \\

    \rowcolor{outdoorgray}
    gate 03 &
    1.127 & 0.699 &
    0.416 & 0.275 &
    \underline{0.218} & \underline{0.137} &
    0.222 & 0.142 &
    0.220 & \underline{0.137} &
    \textbf{0.215} & \textbf{0.135} \\

    \rowcolor{outdoorgray}
    circle 01 &
    \multicolumn{2}{c|}{FAIL} &
    1.748 & 0.507 &
    1.340 & 0.247 &
    1.554 & \textbf{0.166} &
    \textbf{1.329} & \underline{0.243} &
    \underline{1.332} & \textbf{0.166} \\

    \rowcolor{outdoorgray}
    circle 02 &
    \multicolumn{2}{c|}{FAIL} &
    2.579 & 0.525 &
    1.241 & 0.178 &
    \underline{1.247} & \underline{0.175} &
    \textbf{1.238} & \underline{0.175} &
    \underline{1.241} & \textbf{0.159} \\

    \rowcolor{outdoorgray}
    street 01 &
    12.546 & 0.465 &
    2.483 & 0.274 &
    \textbf{0.297} & \textbf{0.136} &
    0.653 & 1.182 &
    0.305 & \textbf{0.136} &
    \underline{0.302} & \textbf{0.136} \\

    \rowcolor{outdoorgray}
    street 02 &
    12.739 & 0.525 &
    4.565 & 0.444 &
    2.878 & \underline{0.222} &
    3.014 & \textbf{0.220} &
    \underline{2.844} & \underline{0.222} &
    \textbf{2.676} & \underline{0.222} \\

    \rowcolor{outdoorgray}
    street 03 &
    \textbf{0.135} & 0.603 &
    \underline{0.146} & 0.507 &
    0.183 & \underline{0.171} &
    0.182 & \underline{0.171} &
    0.183 & \underline{0.171} &
    0.180 & \textbf{0.154} \\

    \rowcolor{outdoorgray}
    street 04 &
    3.155 & 0.445 &
    6.628 & 0.358 &
    0.448 & \underline{0.180} &
    0.623 & 0.190 &
    \underline{0.447} & \textbf{0.179} &
    \textbf{0.442} & \textbf{0.179} \\

    hall 01 &
    \multicolumn{2}{c|}{FAIL} &
    \textbf{0.219} & 1.872 &
    0.292 & \textbf{0.920} &
    0.293 & \textbf{0.920} &
    0.289 & \textbf{0.920} &
    \underline{0.287} & \textbf{0.920} \\

    hall 02 &
    \multicolumn{2}{c|}{FAIL} &
    \textbf{0.285} & 1.700 &
    \underline{0.522} & 0.821 &
    \underline{0.522} & \underline{0.820} &
    0.532 & 0.821 &
    0.524 & \textbf{0.819} \\

    hall 03 &
    \multicolumn{2}{c|}{FAIL} &
    \textbf{0.325} & 2.115 &
    0.571 & 1.297 &
    0.572 & 1.297 &
    0.574 & \underline{1.293} &
    \underline{0.570} & \textbf{1.291} \\

    hall 04 &
    \multicolumn{2}{c|}{FAIL} &
    \textbf{0.937} & 1.046 &
    \underline{1.070} & \underline{0.647} &
    \underline{1.070} & \textbf{0.646} &
    \underline{1.070} & \underline{0.647} &
    \underline{1.070} & \textbf{0.646} \\

    hall 05 &
    \multicolumn{2}{c|}{FAIL} &
    1.897 & 2.145 &
    1.187 & 1.088 &
    \underline{1.186} & \underline{1.087} &
    1.201 & 1.089 &
    \textbf{1.120} & \textbf{1.083} \\

    room 01 &
    \multicolumn{2}{c|}{FAIL} &
    0.558 & 0.100 &
    \underline{0.403} & 0.075 &
    0.404 & \textbf{0.071} &
    0.404 & \underline{0.074} &
    \textbf{0.400} & \underline{0.074} \\

    room 02 &
    \multicolumn{2}{c|}{FAIL} &
    0.330 & 0.109 &
    0.323 & 0.076 &
    0.325 & \underline{0.075} &
    \underline{0.322} & 0.076 &
    \textbf{0.320} & \textbf{0.074} \\

    room 03 &
    \multicolumn{2}{c|}{FAIL} &
    0.563 & 0.117 &
    0.430 & 0.086 &
    \underline{0.429} & \underline{0.085}&
    0.430 & 0.086 &
    \textbf{0.426} & \textbf{0.083} \\

    door 01 &
    1.316 & \textbf{0.278} &
    0.257 & 0.929 &
    0.195 & 0.546 &
    0.462 & 0.551 &
    \underline{0.194} & \underline{0.545} &
    \textbf{0.178} & 0.470 \\

    door 02 &
    \textbf{0.221} & \textbf{0.220} &
    0.223 & 0.754 &
    0.345 & 0.425 &
    0.345 & \underline{0.423} &
    0.345 & \underline{0.423} &
    \underline{0.337} & \underline{0.423} \\

    \bottomrule
  \end{tabular}%
  }
  \end{threeparttable}
\end{table*}

Table~\ref{tab:m2dgr_accuracy} presents the localization results across different sequences. The \textbf{GR-LIO (FP)} generally exhibits degraded localization accuracy, particularly in outdoor sequences. This is because fixed ground plane cannot accommodate changes in the local ground geometry caused by uneven terrain and varying slopes, resulting in increasingly inconsistent ground constraints as the robot moves. In contrast, the proposed B--G parameterization represents the local ground geometry relative to the robot body frame, allowing the ground plane to adapt to changes in the robot pose and local terrain. Even without the planar motion update, \textbf{GR-LIO (w/o PU)} therefore achieves improved localization accuracy over the fixed-plane variant and outperforms FAST-LIO2 on several sequences. With the planar motion update enabled, GR-LIO further improves the estimation accuracy and achieves the best performance in most indoor sequences. In addition, the proposed B--G model enables ground point-to-plane constraints to be constructed without conventional nearest-neighbor searches, allowing more ground observations to be efficiently incorporated into the filter update while reducing the computational cost, as demonstrated in Section~\ref{exp:time}.

LINS and LIO-SAM also achieve competitive localization accuracy on several sequences. However, both methods rely on explicit extraction of edge and planar features, which introduces additional computational overhead. In contrast, GR-LIO directly exploits local ground geometry within the FAST-LIO2 framework, achieving improved localization accuracy while maintaining superior computational efficiency.

\begin{figure}
    \centering
    \includegraphics[width= 0.9\linewidth]{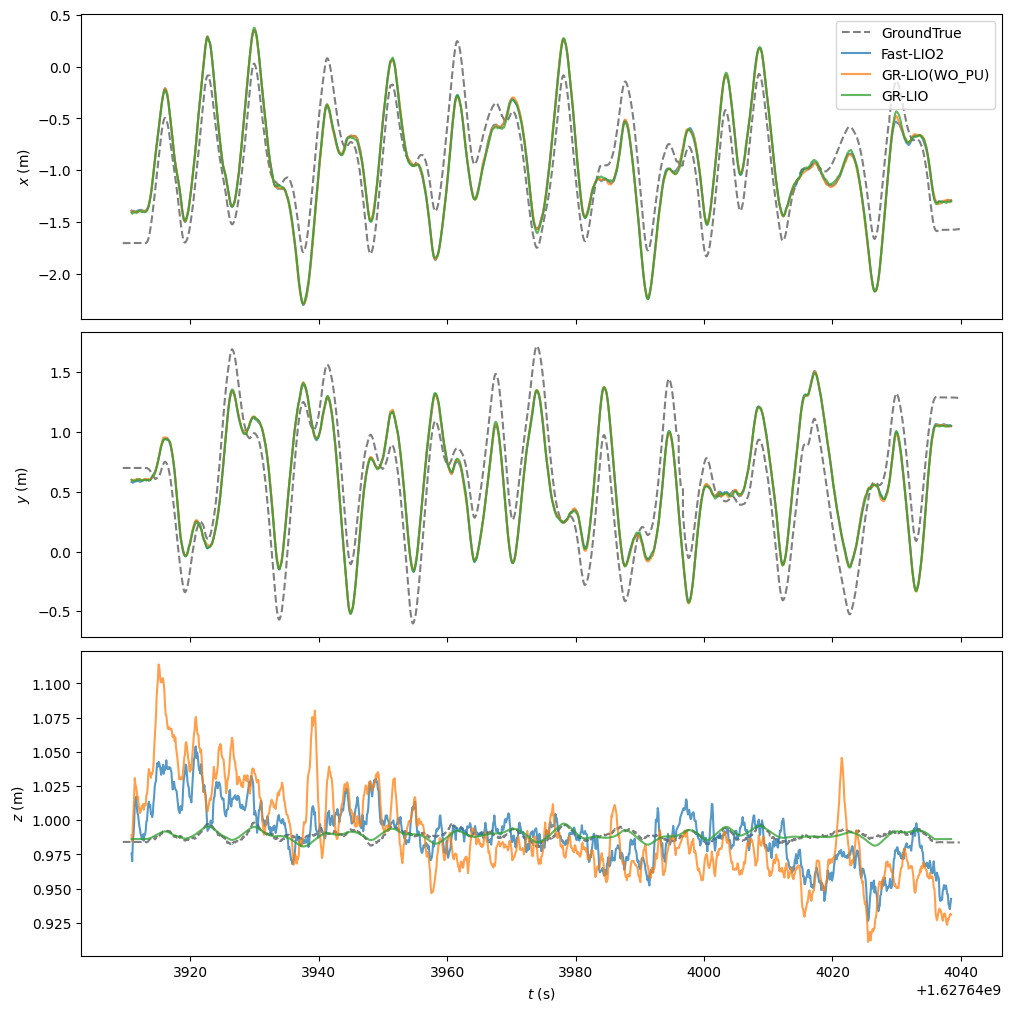}
    \caption{Trajectory comparison of FAST-LIO2 (green) and our GR-LIO (blue) with the ground truth (black dashed) in room 03 along the $x$, $y$, and $z$ axes.}
    \label{fig:reference}
\end{figure}

Fig.~\ref{fig:reference} shows the estimation results of GR-LIO and FAST-LIO2 compared with the ground truth along different axes. It can be observed that the two methods achieve similar performance along the $x$- and $y$-axes. However, along the $z$-axis, GR-LIO remains consistently closer to the ground truth, whereas FAST-LIO2 exhibits noticeable oscillations with an error of approximately 0.02~m. This improvement can be attributed to the planar motion constraint incorporated in GR-LIO, which effectively suppresses drift in the vertical direction.

\begin{figure*}
    \subfloat[gate 01]{\includegraphics[width=0.33\linewidth]{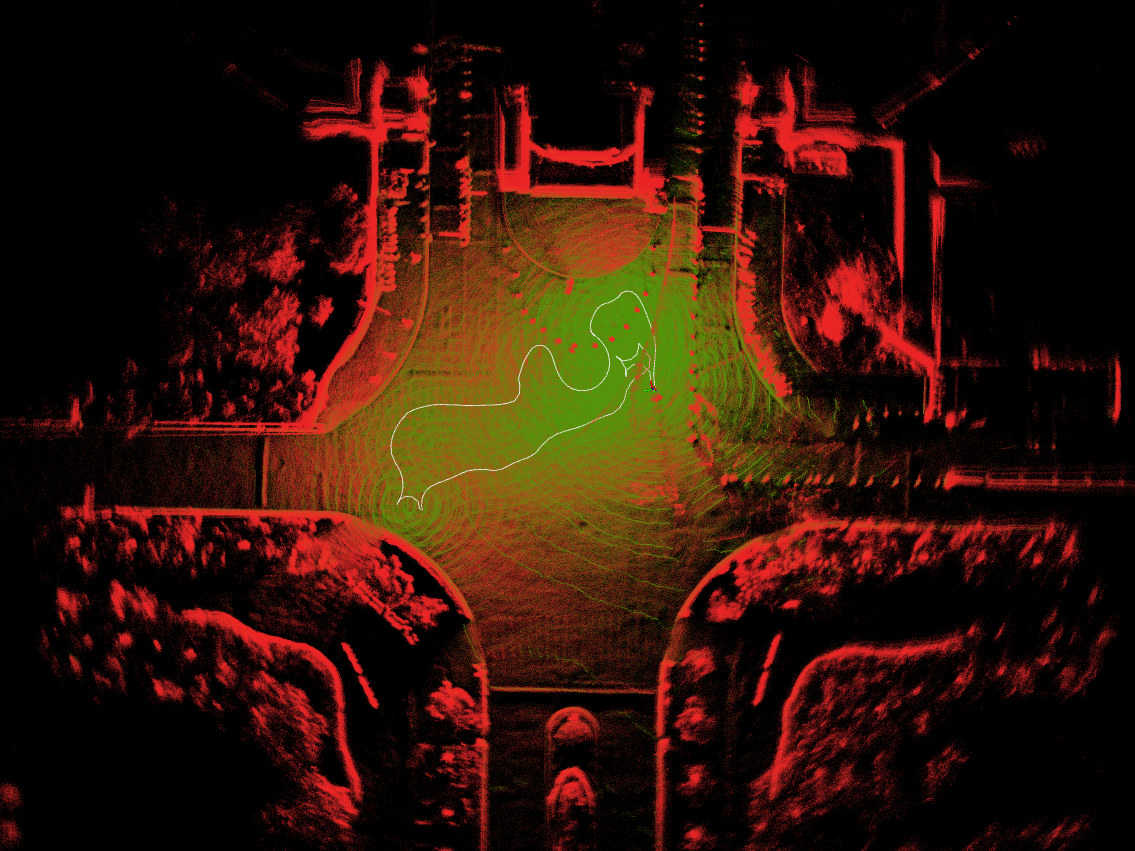}}
    \hfill
    \subfloat[gate 02]{\includegraphics[width=0.33\linewidth]{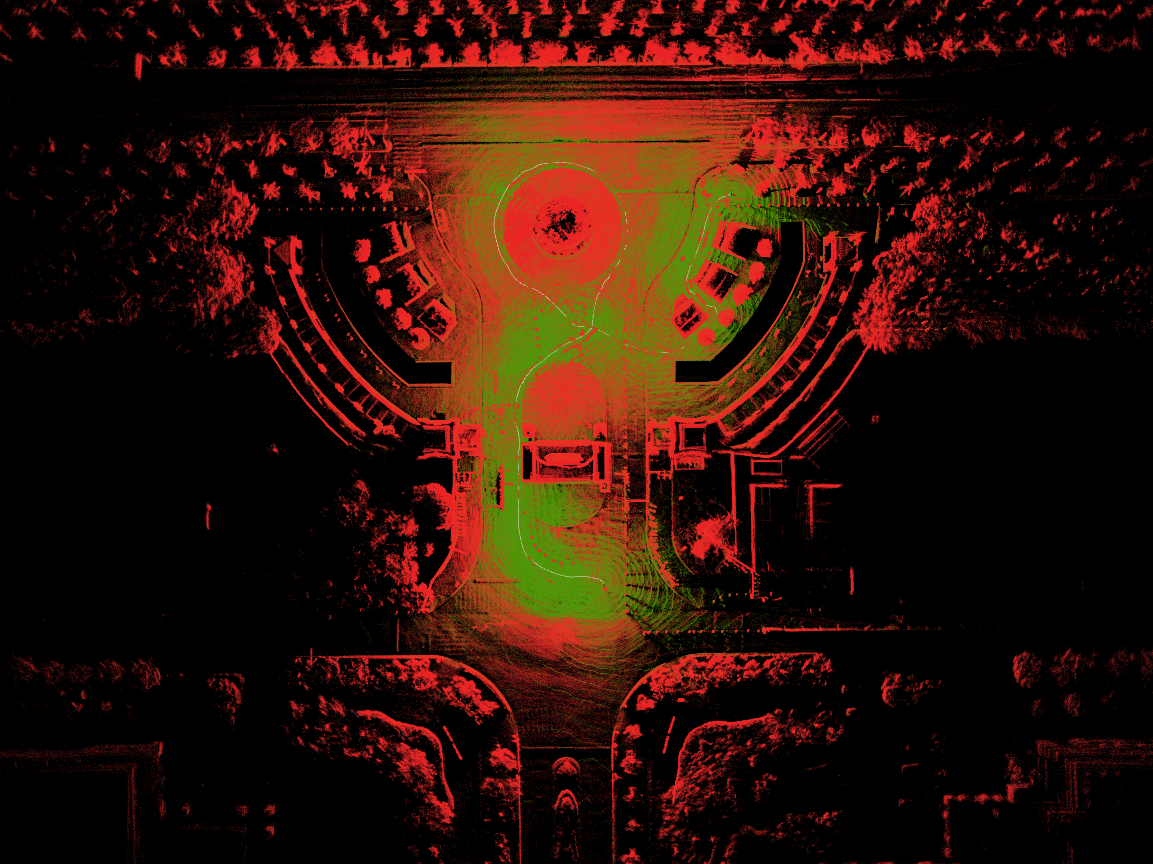}} 
    \hfill
    \subfloat[gate 03]{\includegraphics[width=0.33\linewidth]{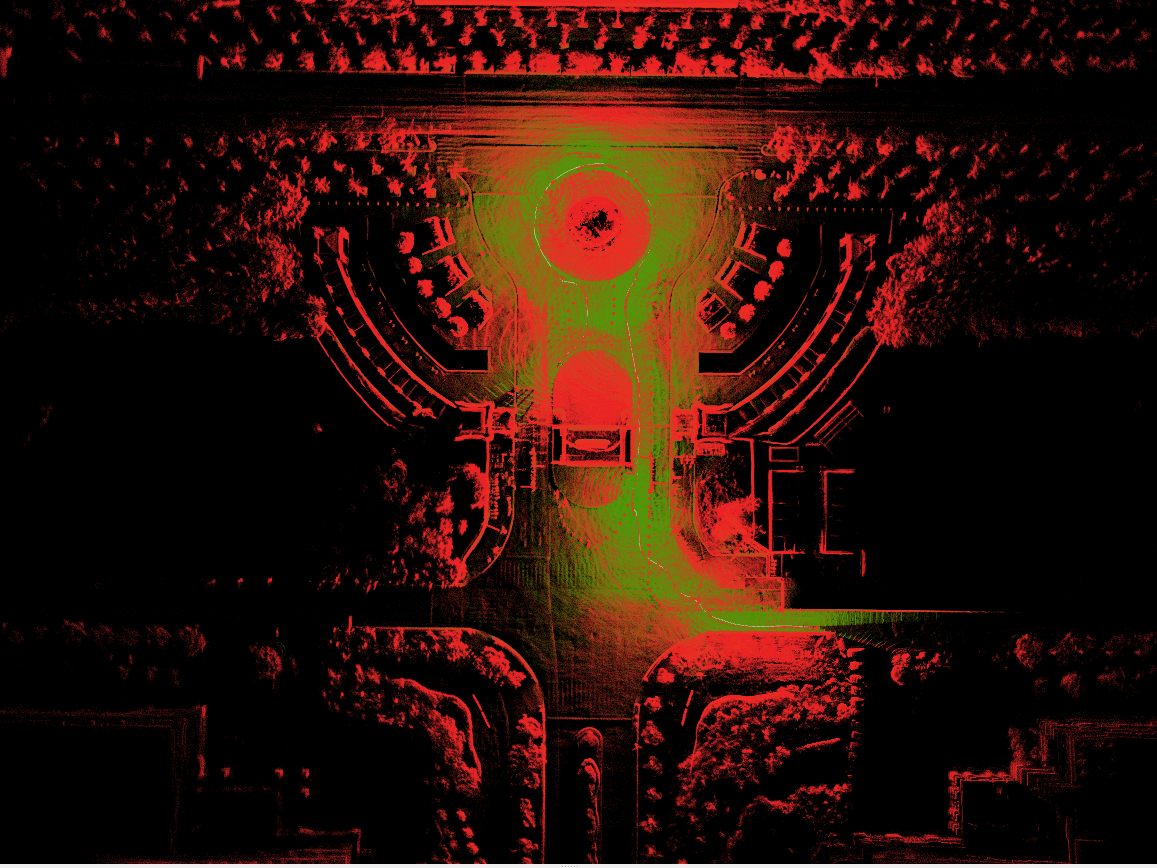}}
    \\
    \subfloat[Street 01]{\includegraphics[width=0.33\linewidth]{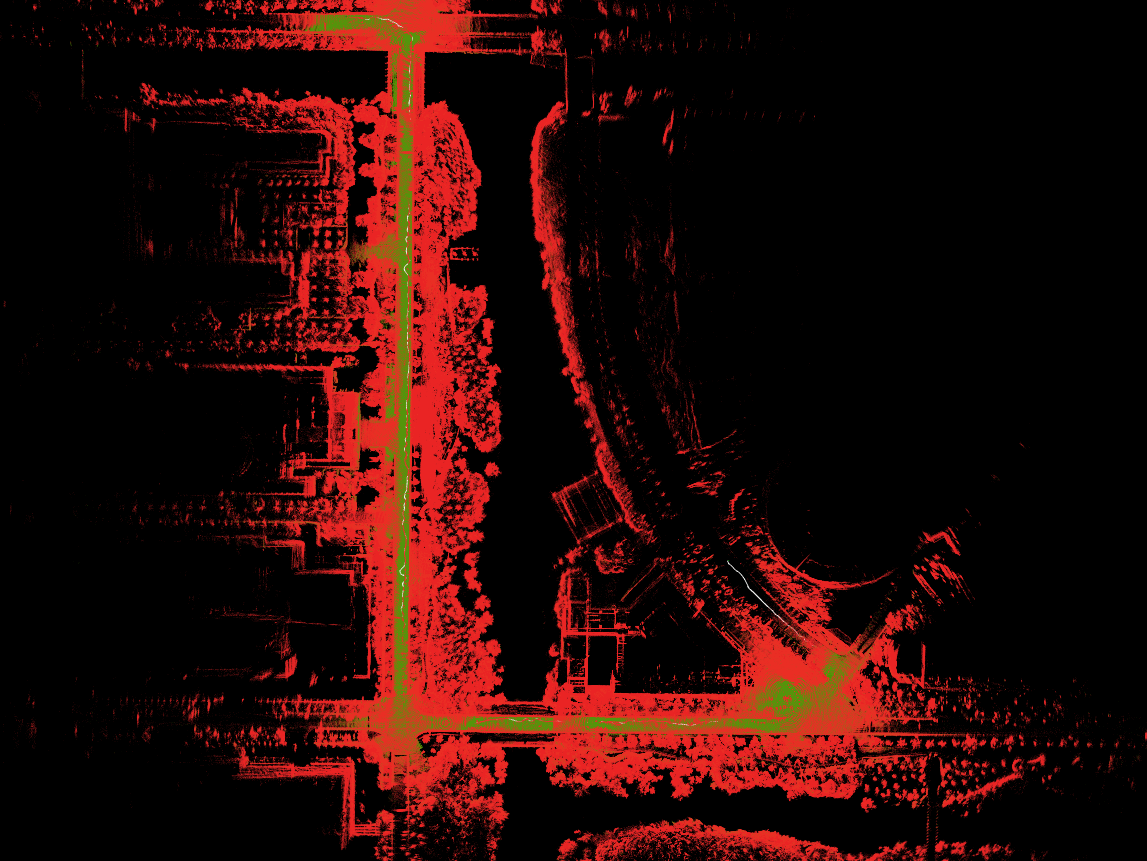}}
    \hfill
    \subfloat[Street 02]{\includegraphics[width=0.33\linewidth]{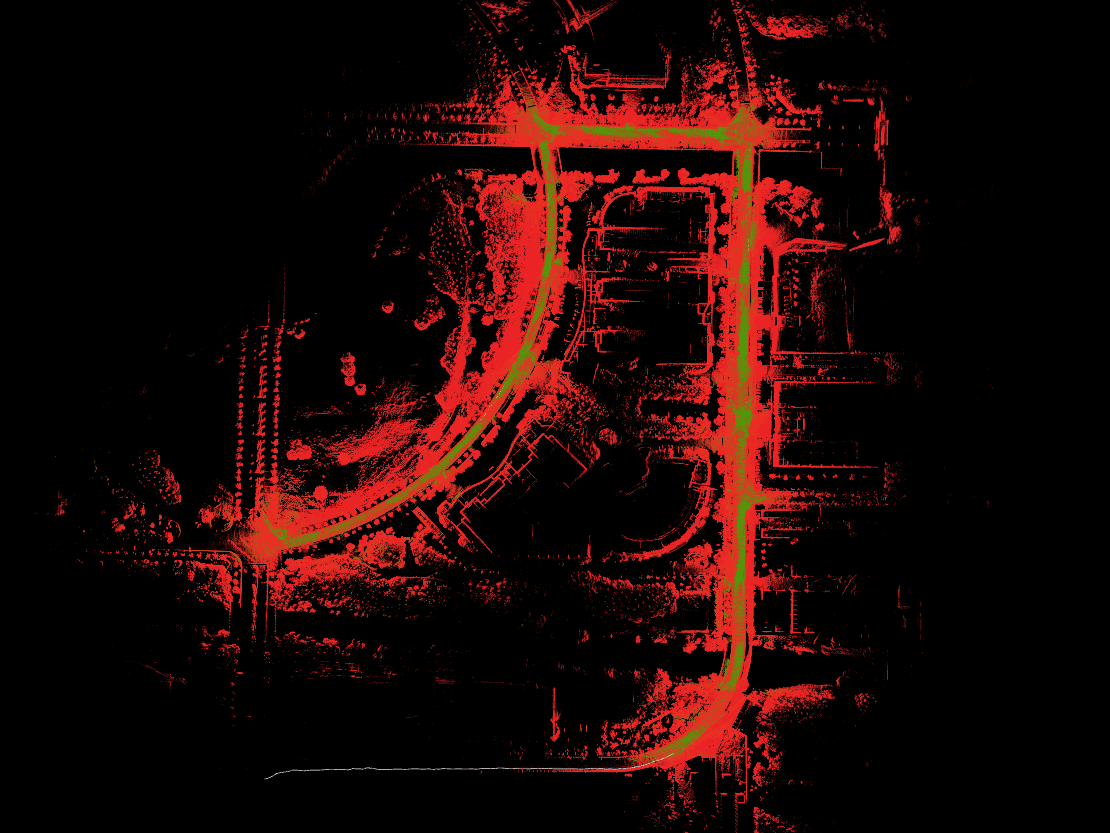}}
    \hfill
    \subfloat[Street 03]{\includegraphics[width=0.33\linewidth]{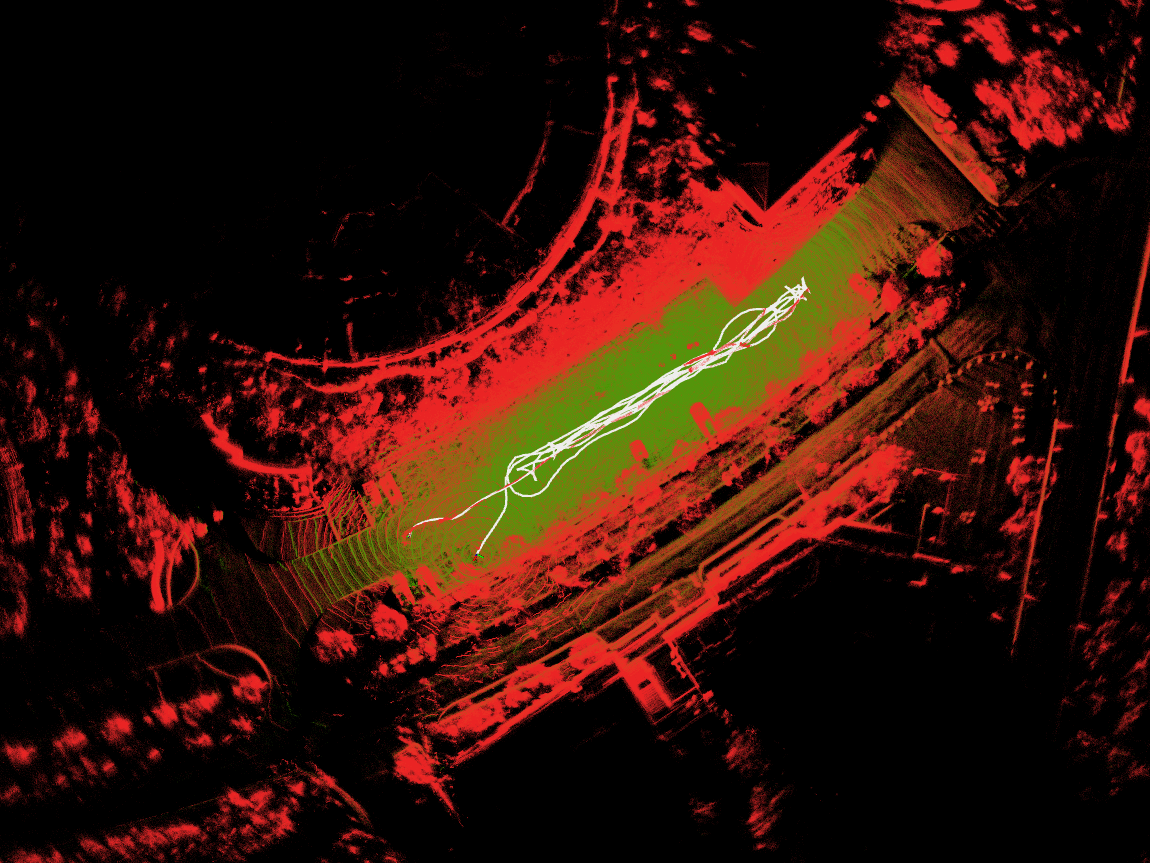}}
    \\
    \subfloat[hall 01]{\includegraphics[width=0.33\linewidth]{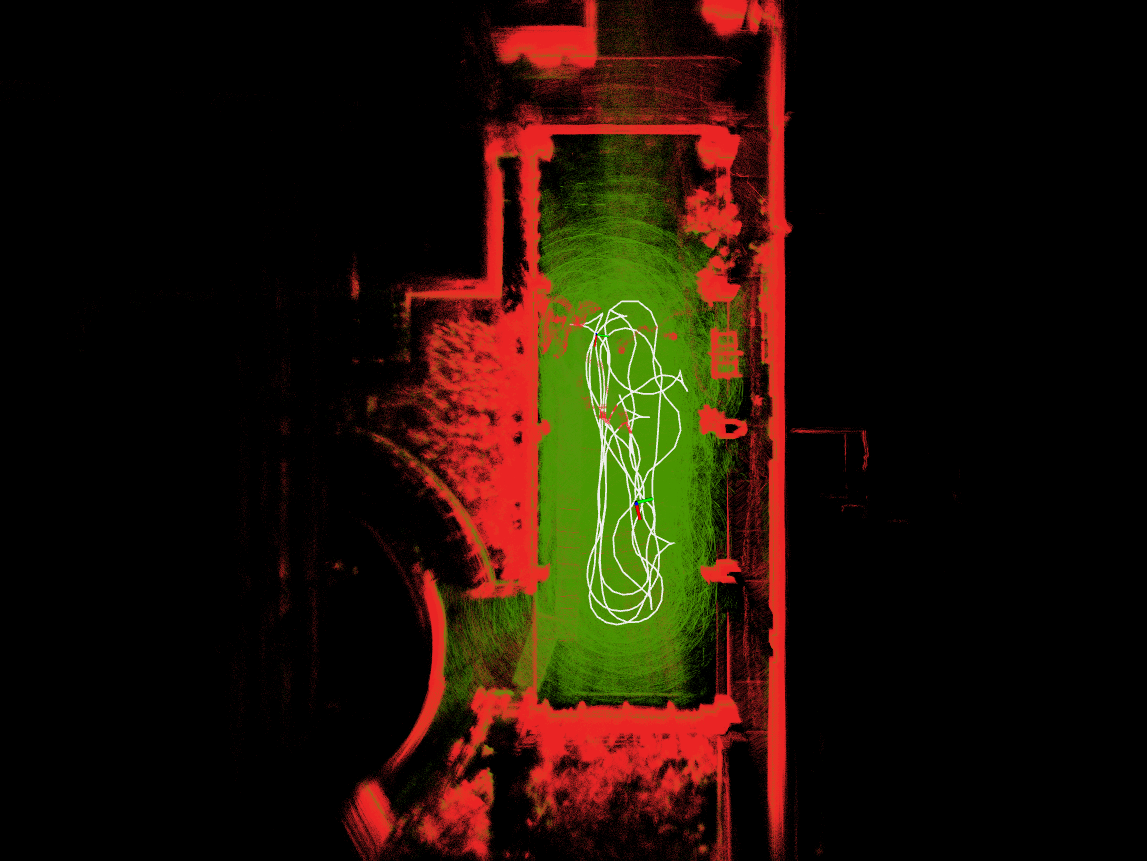}}
    \hfill
    \subfloat[room 03]{\includegraphics[width=0.33\linewidth]{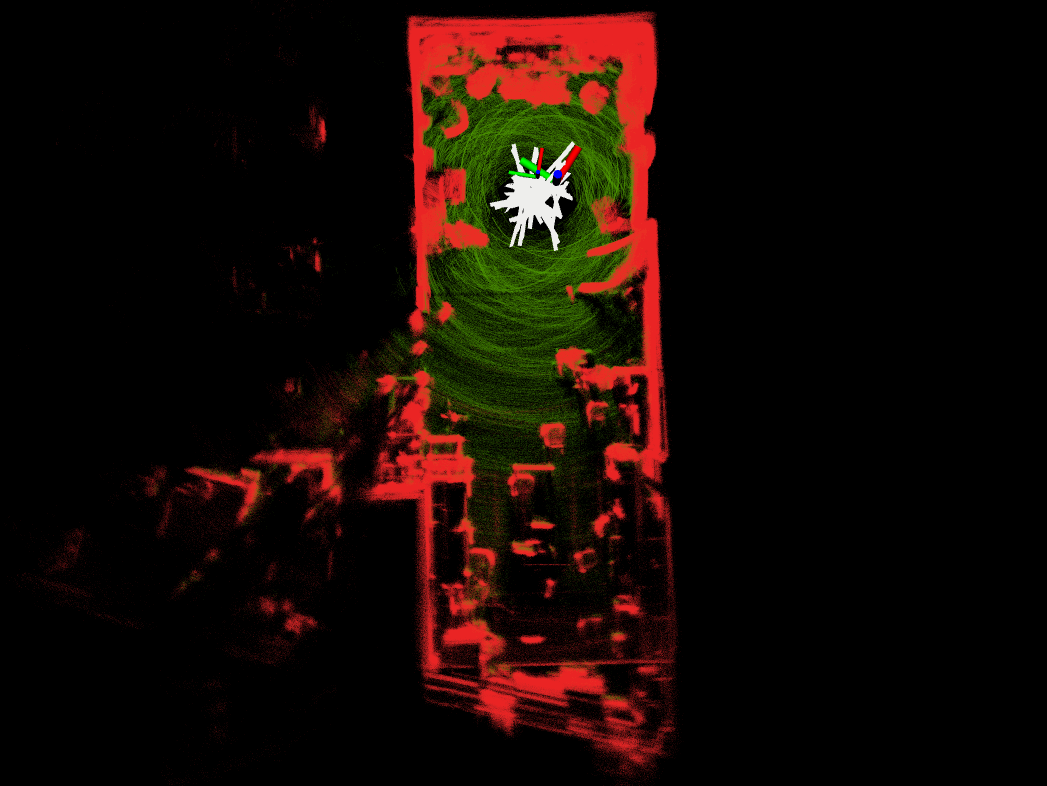}}
    \hfill
    \subfloat[door 02]{\includegraphics[width=0.33\linewidth]{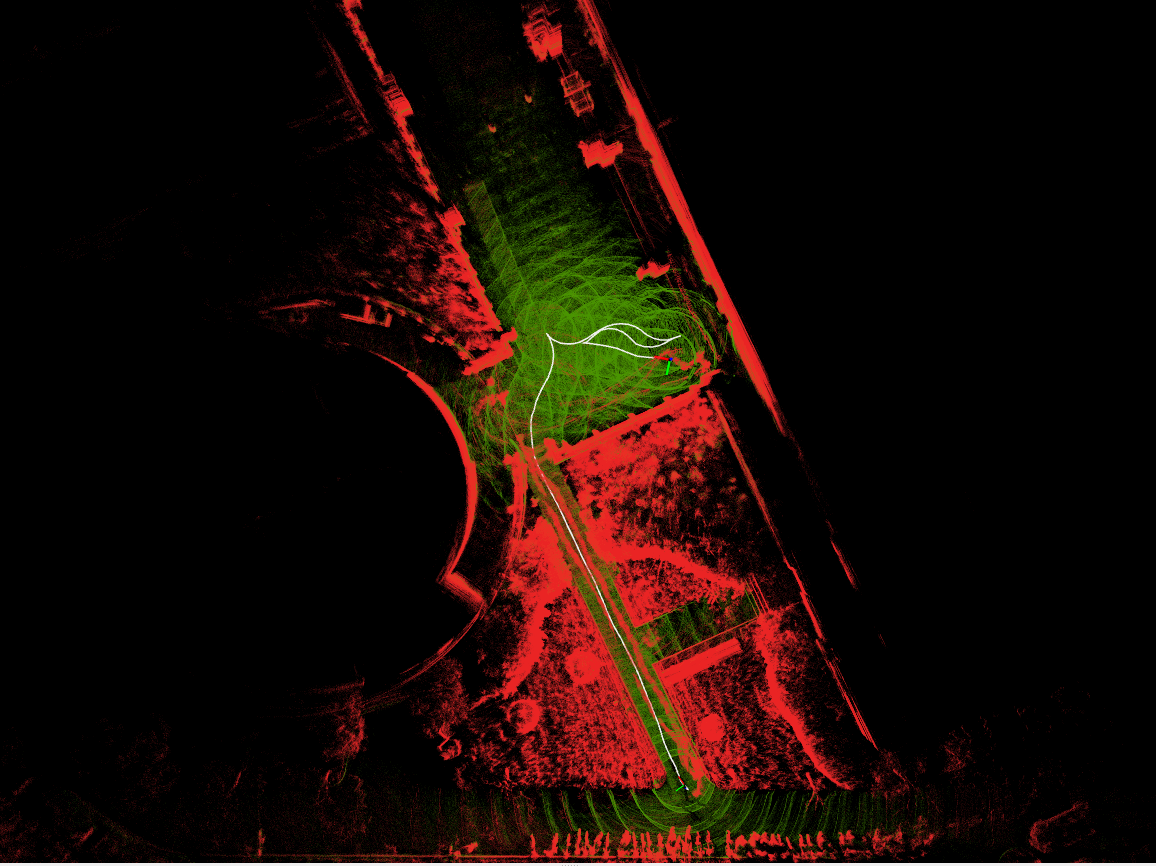}}  
    \caption{Representative mapping results and estimated trajectories on the M2DGR dataset. The white curves represent the estimated robot trajectories. Green points indicate segmented local ground points, while red points denote non-ground points.}
    \label{fig:m2dgr_mapping}
\end{figure*}

Fig.~\ref{fig:m2dgr_mapping} shows representative segmented mapping results and estimated trajectories produced by the proposed GR-LIO framework. The segmented local ground points are highlighted in green, while the remaining non-ground points are shown in red. The reconstructed maps accurately preserve the structural layout of the real environments. Moreover, the proposed method consistently segments local ground points across diverse environments, providing reliable geometric observations for state estimation.

\subsubsection{M3DGR Datasets}
Compared with M2DGR, the solid-state LiDAR used in this dataset has a smaller FOV and observes fewer ground points due to its sensing characteristics, resulting in many ground points being unable to establish reliable constraints in conventional methods. The localization accuracy results are summarized in Table~\ref{tab:m3dgr_accuracy}.

\begin{table}[t]
  \centering
  \caption{RMSE comparison of FAST-LIO2 and GR-LIO variants on the M3DGR dataset
  (ATE$_t$: m, RPE$_t$: m).}
  \label{tab:m3dgr_accuracy}
  \setlength{\tabcolsep}{3.5pt}
  \renewcommand{\arraystretch}{1.15}

  \resizebox{\columnwidth}{!}{%
  \begin{tabular}{c|cc|cc|cc|cc}
    \toprule
    \multirow{2}{*}{\textbf{Sequence}} &
    \multicolumn{2}{c|}{\textbf{FAST-LIO2~\cite{c4}}} &
    \multicolumn{2}{c|}{\textbf{GR-LIO (FP)}} &
    \multicolumn{2}{c|}{\textbf{GR-LIO(w/o PU)}} &
    \multicolumn{2}{c}{\textbf{GR-LIO}} \\
    
    \cmidrule(lr){2-3}
    \cmidrule(lr){4-5}
    \cmidrule(lr){6-7}
    \cmidrule(lr){8-9}
    
    & ATE$_t$ & RPE$_t$
    & ATE$_t$ & RPE$_t$
    & ATE$_t$ & RPE$_t$
    & ATE$_t$ & RPE$_t$ \\
    \midrule

    \rowcolor{outdoorgray}
    Outdoor01 &
    0.244 & 0.151 &
    17.56 & 12.74 &
    \underline{0.242} & \underline{0.150} &
    \textbf{0.216} &  \textbf{0.142} \\

    \rowcolor{outdoorgray}
    Outdoor04 &
    \underline{0.826} & \underline{0.155}  &
    0.879 & 0.276 &
    0.910 & \underline{0.155}  &
    \textbf{0.640} & \textbf{0.127} \\

    \rowcolor{outdoorgray}
    Dynamic03 &
    1.144 & 0.144  &
    \textbf{0.044} & \textbf{0.008} &
    0.163 & 0.144  &
    \underline{0.138} & \underline{0.122} \\

    \rowcolor{outdoorgray}
    Dynamic04 &
    0.297 & 0.144 &
    0.761 & 0.477 &
    \underline{0.273} & \underline{0.143} &
    \textbf{0.123} & \textbf{0.136} \\

    \rowcolor{outdoorgray}
    Varying-illu03 &
    \underline{0.638} & \underline{0.166} &
    0.812 & 0.168 &
    0.654 & 0.171 &
    \textbf{0.396} & \textbf{0.158} \\

    \rowcolor{outdoorgray}
    Varying-illu04 &
    0.439 & \underline{0.160} &
    0.998 & 0.161 &
    \textbf{0.420} & \textbf{0.159} &
    \underline{0.422} & \textbf{0.159} \\

    \rowcolor{outdoorgray}
    Varying-illu05 &
    0.218 & \textbf{0.094} &
    0.588 & 0.095 &
    \underline{0.218} & \textbf{0.094} &
    \textbf{0.217} & \textbf{0.094} \\

    \rowcolor{outdoorgray}
    Dark01 &
    0.214 & 0.098 &
    0.291 & \underline{0.098} &
    \textbf{0.192} & 0.100 &
    \underline{0.206} & \textbf{0.096} \\

    \rowcolor{outdoorgray}
    Dark02 &
    0.786 &  0.134 &
    1.216 &\underline{0.126} &
    \underline{0.731} &  0.134 &
    \textbf{0.536} & \textbf{0.122} \\

    \rowcolor{outdoorgray}
    GNSS-denial01 &
    7.90 & 0.14 &
    12.43 & 0.15 &
    \textbf{5.95} & \textbf{0.12} &
    6.59 & 0.15 \\

    \rowcolor{outdoorgray}
    GNSS-denial02 &
    \underline{0.65} & \textbf{0.47} &
    2.56 & 0.5&
    \underline{0.66} & \textbf{0.47} &
    \textbf{0.58} & \textbf{0.47} \\

    \bottomrule
  \end{tabular}%
  }
\end{table}

Overall, \textbf{GR-LIO (FP)} shows the worst performance among the evaluated variants. Since the M3DGR sequences contain long outdoor trajectories with more complex and varying road conditions, the ground plane estimated during initialization can become inconsistent with the actual local ground geometry. In contrast, the proposed B--G parameterization adapts the ground model to the changing terrain and provides more reliable ground constraints. As a result, GR-LIO without the planar motion update outperforms FAST-LIO2 on 8 out of 11 sequences, with further improvements achieved after incorporating the planar motion update, which provides additional constraints on the vertical and rotational degrees of freedom.
\begin{figure}
    \centering
    \includegraphics[width= 0.9 \linewidth]{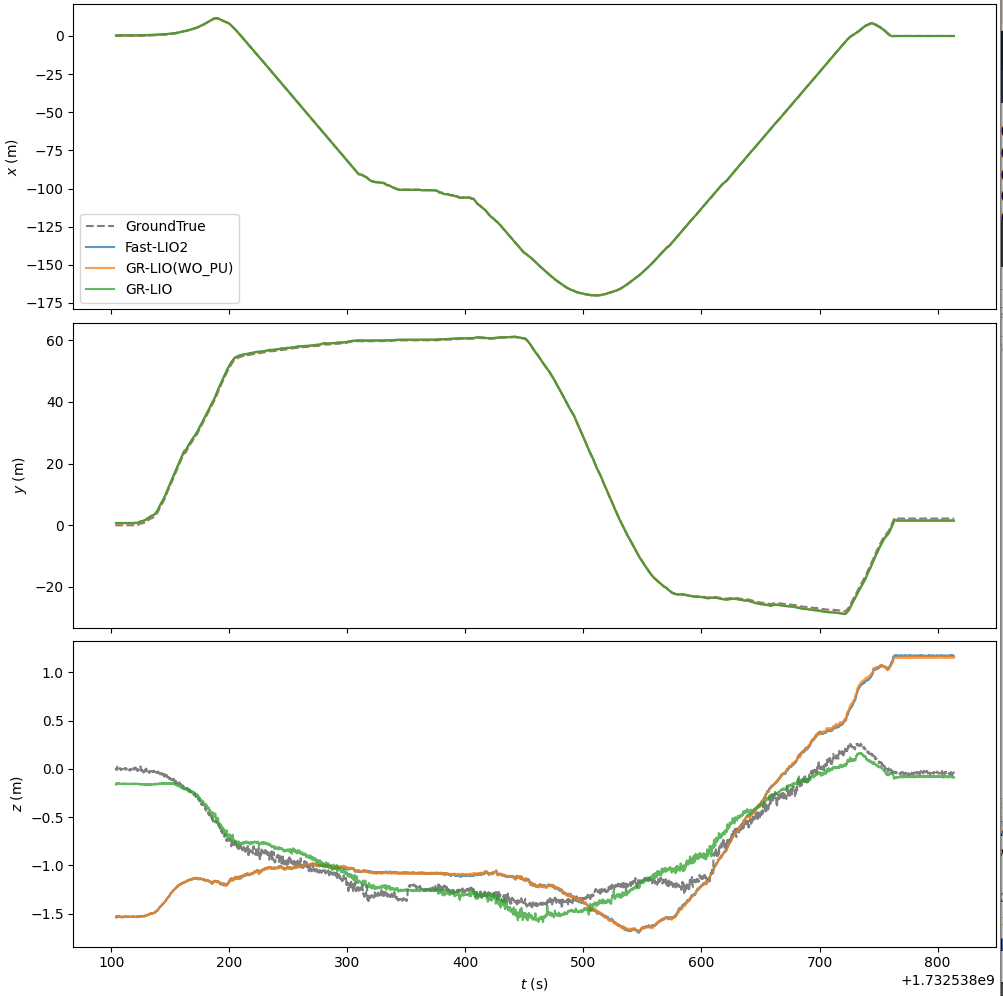}
    \caption{Comparison of FAST-LIO2 (blue), GR-LIO without planar motion update (orange), and GR-LIO (green) against the ground truth (black dashed) along the $x$, $y$, and $z$ axes in M3DGR Dark02.}
    \label{fig:M3DGR}
\end{figure}

Fig.~\ref{fig:M3DGR} compares FAST-LIO2, GR-LIO without planar motion update (\textbf{w/o PU}), and GR-LIO against the ground truth along the three coordinate axes for the Dark02 sequence. All methods exhibit comparable performance along the $x$- and $y$-axes. Moreover, FAST-LIO2 and GR-LIO \textbf{w/o PU} achieve similar accuracy throughout the sequence, indicating that the incorporation of ground plane priors does not degrade estimation accuracy. However, significant differences can be observed along the $z$-axis. FAST-LIO2 and GR-LIO \textbf{w/o PU} suffer from noticeable oscillatory errors, reaching up to 1.5~m, whereas our GR-LIO remains consistently closer to the ground truth, with a maximum error below 0.1~m. These results demonstrate that the proposed planar motion update effectively suppresses vertical drift and thereby improves overall localization accuracy for ground robot applications.

\begin{table}[t]
  \centering
    \caption{RMSE comparison under different B-G height configurations on the M2DGR dataset (APE$_t$: m, APE$_r$: \protect\textdegree, RPE$_t$: m). \\ The three configurations correspond to initialization only, online calibration, and ground-truth B-G height.}
  \label{tab:onlineCarlib}

  \setlength{\tabcolsep}{3pt}
  \renewcommand{\arraystretch}{1.1}

  \resizebox{\columnwidth}{!}{
  \begin{tabular}{c|ccc|ccc|ccc}
    \toprule

    \multirow{2}{*}{\textbf{Sequence}} &
    \multicolumn{3}{c|}{$\mathbf{GR-LIO}_{\mathrm{Init}}$} &
    \multicolumn{3}{c|}{$\mathbf{GR-LIO}_{\mathrm{Calib}}$} &
    \multicolumn{3}{c}{$\mathbf{GR-LIO}_{\mathrm{GT}}$} \\

    \cmidrule(lr){2-4}
    \cmidrule(lr){5-7}
    \cmidrule(lr){8-10}

    & APE$_t$ & APE$_r$ & RPE$_t$
    & APE$_t$ & APE$_r$ & RPE$_t$
    & APE$_t$ & APE$_r$ & RPE$_t$ \\

    \midrule

    \rowcolor{outdoorgray}
    gate 01
    & 0.176 & 2.740 & 0.145
    & 0.172 & 2.740 & 0.144
    & 0.172 & 2.739 & 0.144 \\

    \rowcolor{outdoorgray}
    gate 02
    & 0.341 & 2.732 & 0.185
    & 0.326 & 2.732 & 0.142
    & 0.320 & 2.732 & 0.140 \\

    \rowcolor{outdoorgray}
    street 01
    & 0.707 & 2.749 & 0.137
    & 0.286 & 2.745 & 0.136
    & 0.308 & 2.746 & 0.136 \\

    \rowcolor{outdoorgray}
    street 02
    & 3.259 & 2.744 & 0.223
    & 2.898 & 2.743 & 0.222
    & 2.876 & 2.743 & 0.222 \\

    hall 01
    & 0.294 & 2.828 & 0.920
    & 0.290 & 2.828 & 0.920
    & 0.287 & 2.828 & 0.920 \\

    hall 02
    & 0.532 & 2.828 & 0.820
    & 0.523 & 2.828 & 0.820
    & 0.524 & 2.828 & 0.819 \\

    hall 03
    & 0.580 & 2.828 & 1.300
    & 0.576 & 2.828 & 1.293
    & 0.570 & 2.828 & 1.291 \\

    hall 04
    & 1.078 & 2.828 & 0.647
    & 1.070 & 2.828 & 0.646
    & 1.070 & 2.828 & 0.646 \\

    hall 05
    & 1.188 & 2.828 & 1.086
    & 1.137 & 2.828 & 1.084
    & 1.120 & 2.828 & 1.083 \\

    \bottomrule
  \end{tabular}
  }
\end{table}

\subsection{LIO with Unknown G-B Height}
In this section, we evaluate the proposed system under the practical scenario where the B-G height is unknown or inaccurately initialized. This setting relaxes the assumption that the B-G height is perfectly known and allows us to assess the effectiveness of the proposed initialization and online calibration strategies. Compared with solid-state LiDAR sensors, spinning LiDAR sensors typically provide denser ground observations and therefore offer more informative measurements for B-G height estimation. Consequently, the evaluation is conducted on the M2DGR dataset, which was collected using a spinning LiDAR sensor.

\begin{figure*}[t]
    \centering
    \subfloat[hall 01]{\includegraphics[width=0.45\textwidth]{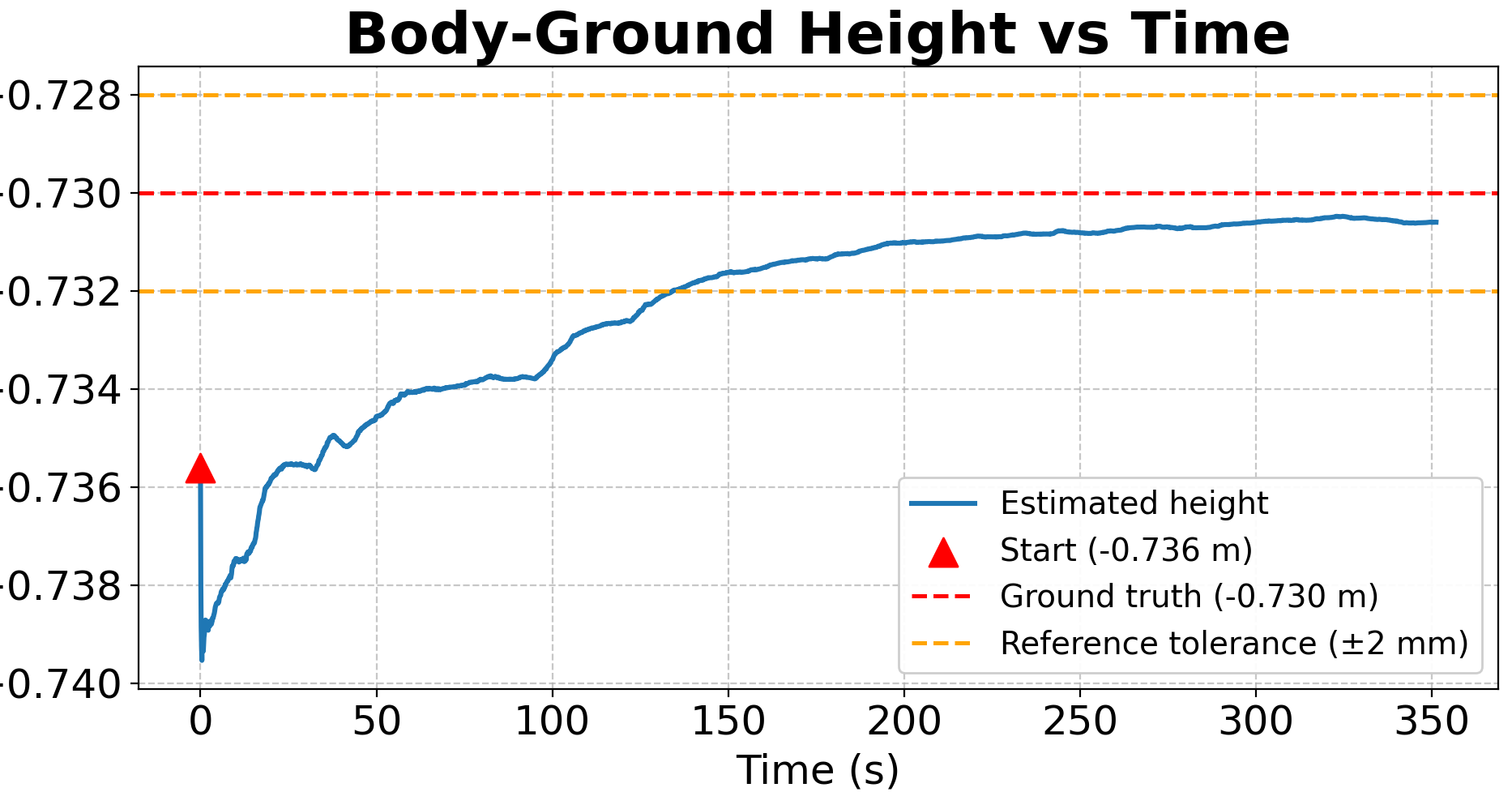}}
    \hspace{0.01\textwidth}
    \subfloat[hall 05]{\includegraphics[width=0.45\textwidth]{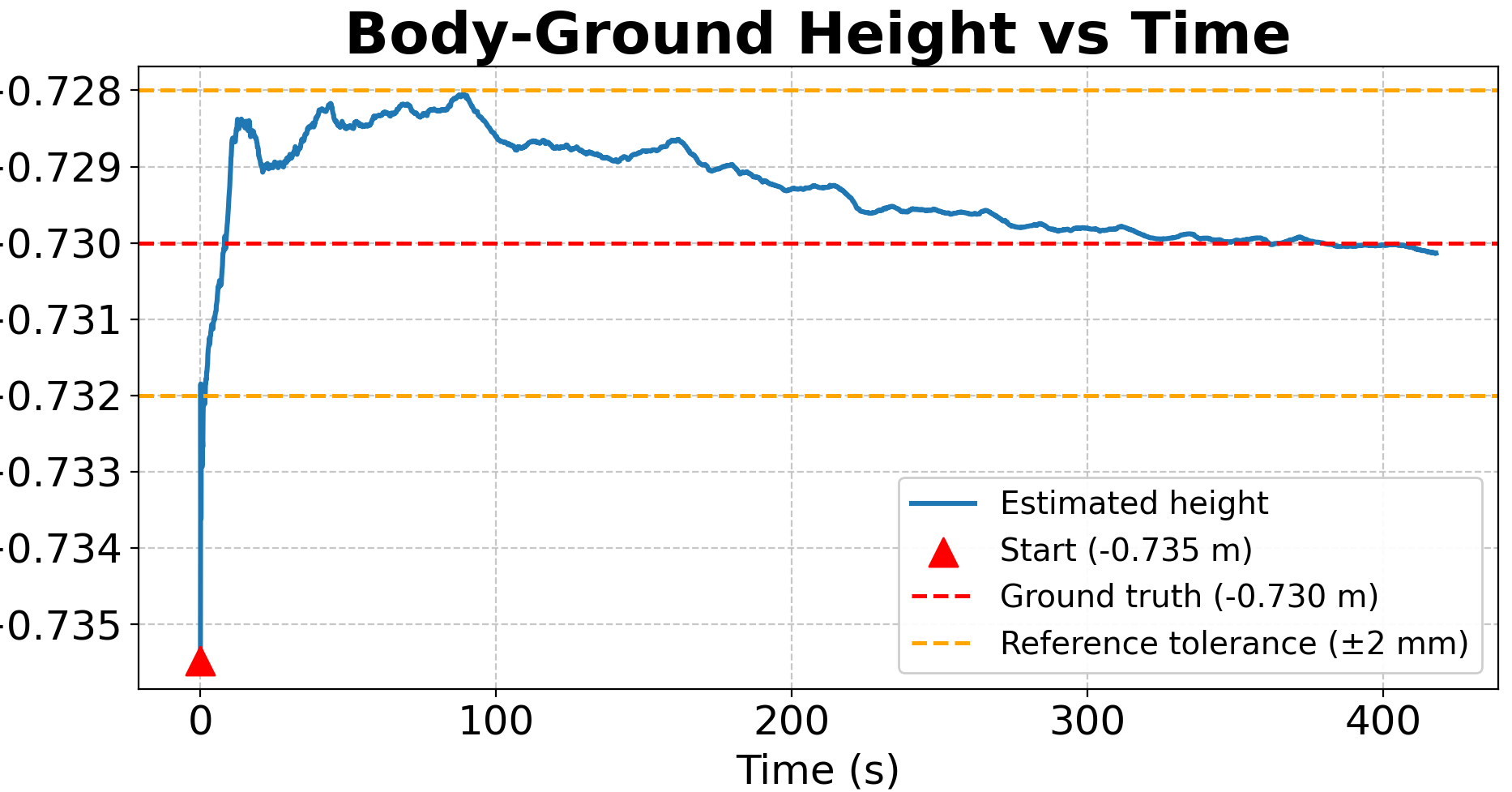}}
    \hspace{0.01\textwidth}
    \subfloat[gate 01]{\includegraphics[width=0.45\textwidth]{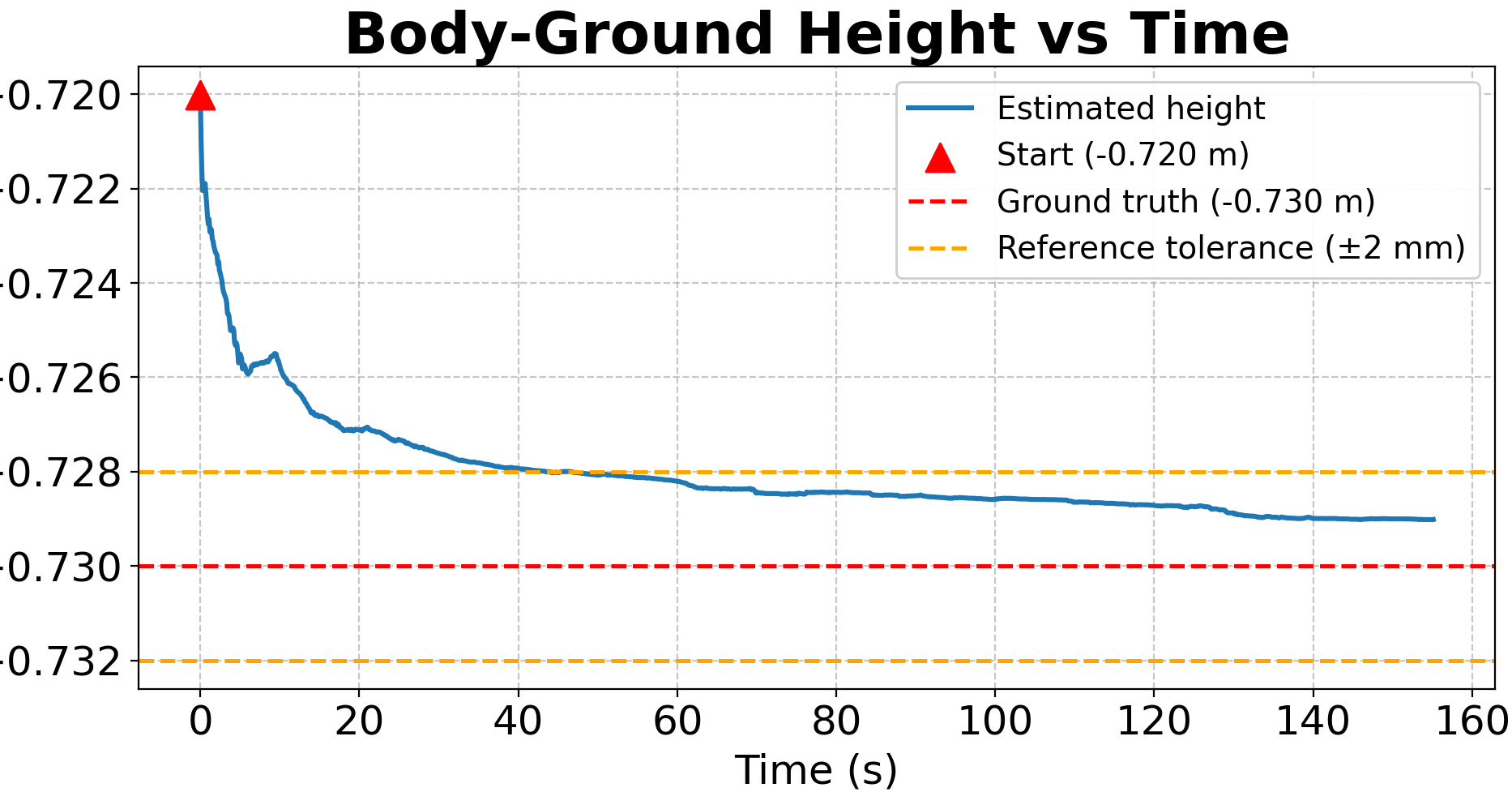}}
    \hspace{0.01\textwidth}
    \subfloat[gate 03]{\includegraphics[width=0.45\textwidth]{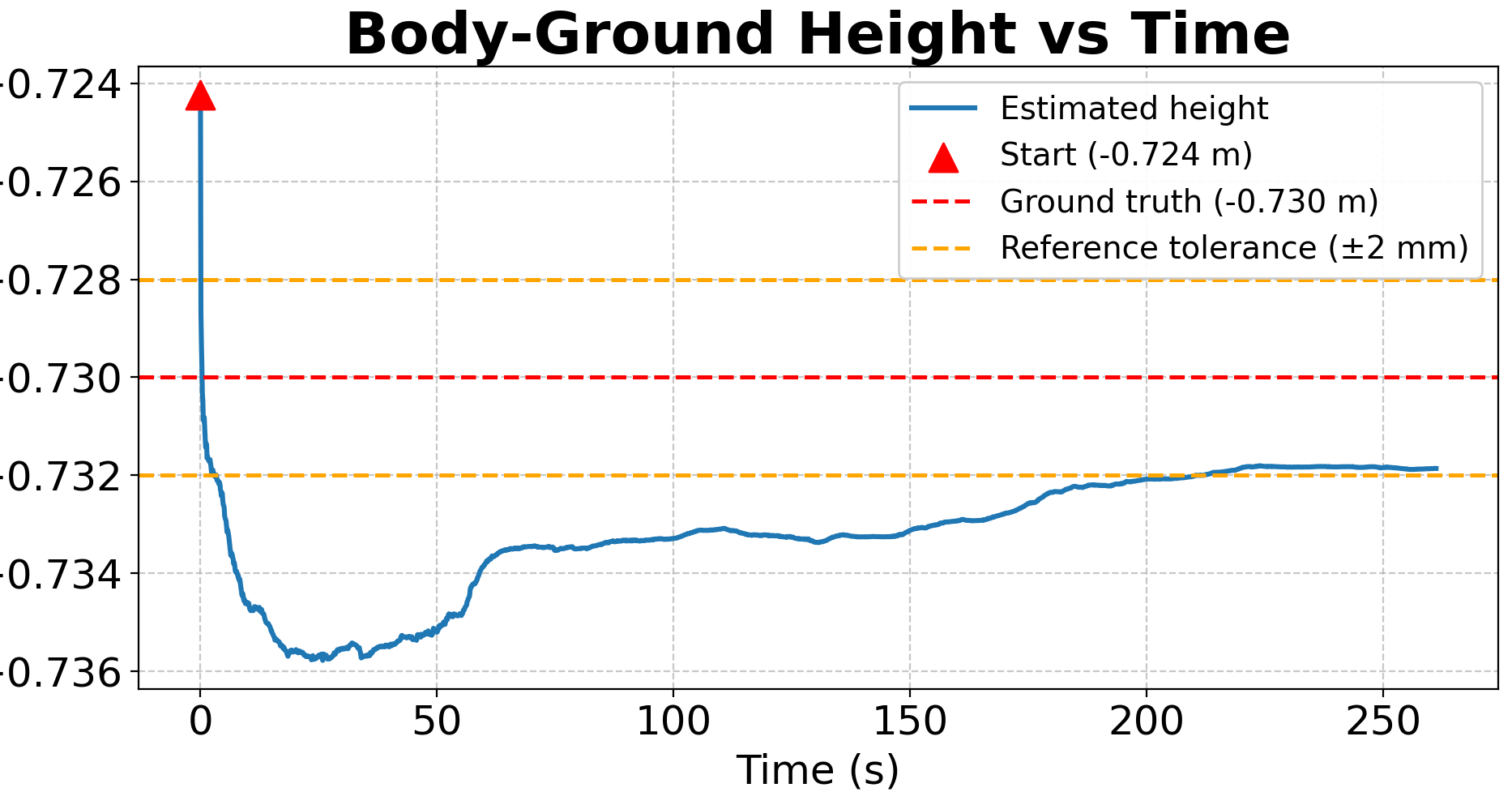}}
    \caption{Initialization and convergence behavior of the ground height estimate during online calibration on the M2DGR dataset. (a)(b) show indoor sequences, while (c)(d) show outdoor sequences. The red triangle indicates the initialization result, and the red dashed line denotes the ground truth.}
    \label{fig:OnlineCalib}
\end{figure*}

\subsubsection{Initialization and Online Calibration Performance}
We first evaluate whether the proposed initialization method can estimate the B-G height from an unknown initial condition and subsequently converge toward the ground truth through online calibration. The method is tested on four representative sequences, and the results are shown in Fig.~\ref{fig:OnlineCalib}.

The proposed initialization method consistently provides an initial B-G height estimate close to the ground truth across different sequences, with an error within $\pm 0.01$~m. During subsequent operation, the online calibration module continuously refines the estimated height, which gradually converges toward the ground truth while the associated covariance decreases. After convergence, the estimation error remains within approximately $\pm 0.002$~m in both indoor and outdoor environments. These results demonstrate the effectiveness of the proposed initialization and online calibration framework and indicate that the estimated B-G height is sufficiently accurate for incorporating ground-plane observations into the state estimation process.

\subsubsection{Online Calibration}
To evaluate the effectiveness of the proposed online B-G height calibration, we compare three different configurations of the proposed system. In the first configuration ($\mathbf{GR-LIO}_{Init}$), the B-G height is initialized using the proposed initialization method and remains fixed throughout the estimation process. In the second configuration ($\mathbf{GR-LIO}_{\mathrm{Calib}}$), the initialized B-G height is further refined through online calibration. As a reference, a third configuration ($\mathbf{GR-LIO}_{GT}$) uses the ground truth B-G height without online calibration, representing the ideal case where the height parameter is perfectly known. The localization results obtained under these three configurations are summarized in Table~\ref{tab:onlineCarlib}.

The results show that enabling online calibration consistently improves localization accuracy across all sequences compared with using only the initialized B-G height. This indicates that the online calibration module effectively compensates for residual initialization errors and refines the estimated B-G height, thereby improving the overall state estimation accuracy. Furthermore, the performance achieved with online calibration is comparable to that obtained using the ground truth B-G height, with only marginal differences in all metrics. These results demonstrate that the proposed online calibration method can accurately estimate the B-G height and achieve localization performance close to the ideal case where the true B-G height is known a prior.

\subsection{Computational Efficiency Analysis}

\label{exp:time}
\begin{figure}
    \centering
    \subfloat[M2DGR room 01 (Indoor)]{\includegraphics[width=0.9\linewidth]{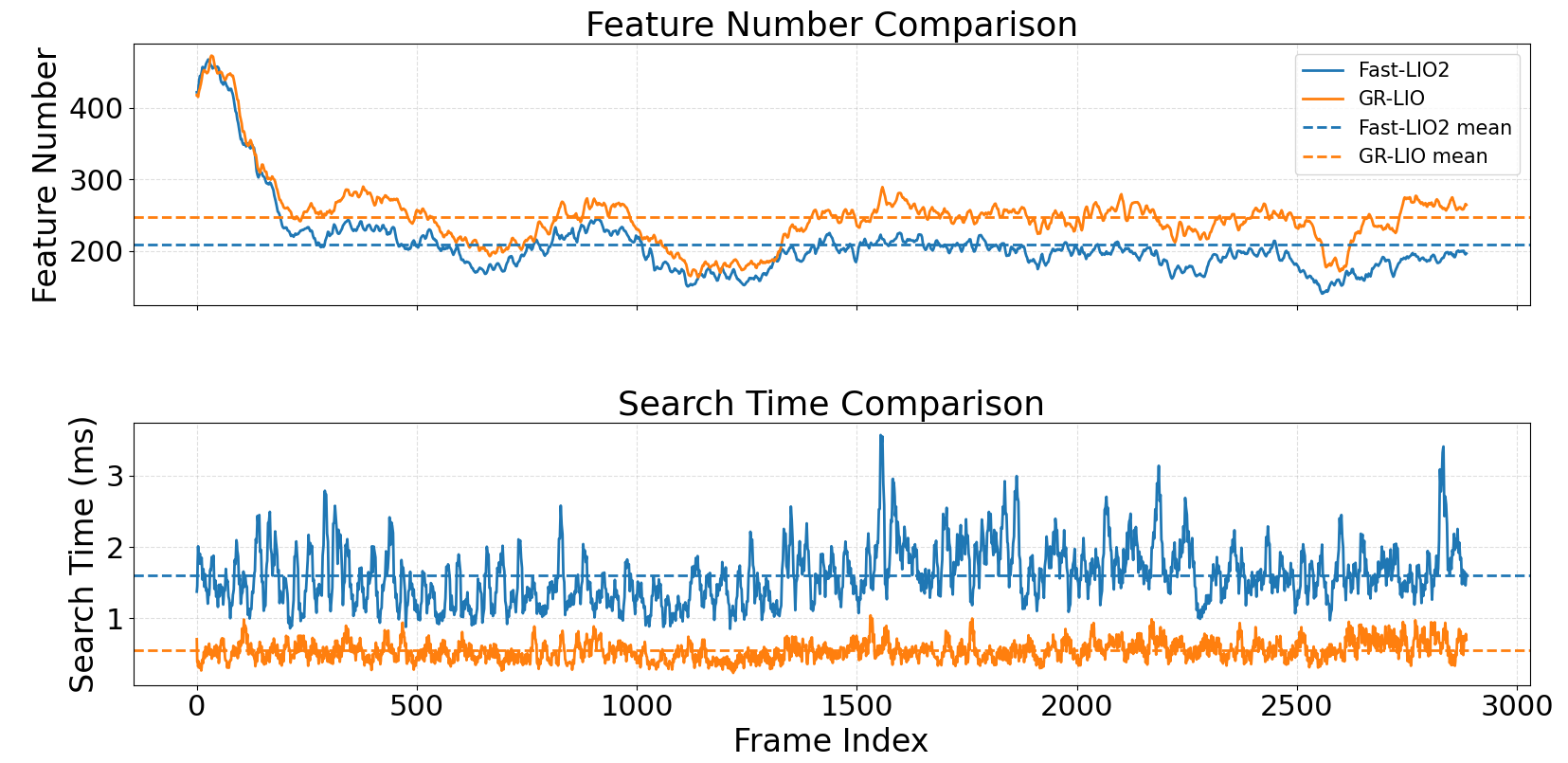}}
    \\
    \subfloat[M2DGR gate 01 (Outdoor)]{\includegraphics[width=0.9\linewidth]{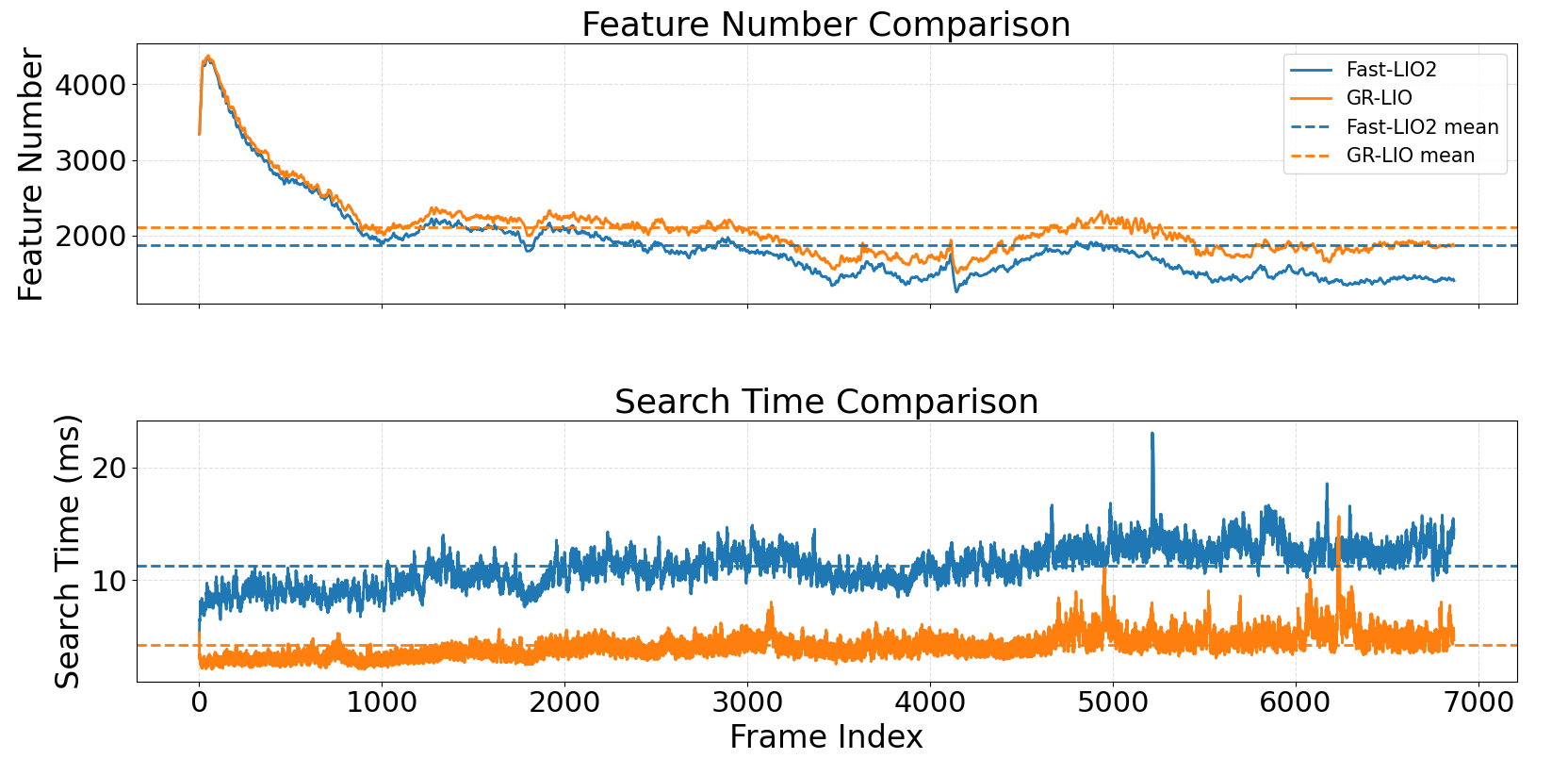}}
    \caption{Comparison of FAST-LIO2 and GR-LIO in terms of the number of features used for state update and the corresponding search time at each frame. The blue and orange curves represent FAST-LIO2 and GR-LIO, respectively, while the dashed lines indicate the corresponding mean values.}
    \label{fig:time_compare}
\end{figure}

\subsubsection{Feature Utilization and Search Time}
To demonstrate the computational efficiency of our GR-LIO, we analyze the number of LiDAR points used for state updates and the corresponding nearest-neighbor search time per frame under two representative scenarios, and compare them with FAST-LIO2. The results are illustrated in Fig.~\ref{fig:time_compare}.

As shown in Fig.~\ref{fig:time_compare}, GR-LIO utilizes more points for state updates while maintaining lower search time than FAST-LIO2. In indoor environments, GR-LIO uses an average of 250 points with a search time of 0.5 ms, compared with approximately 200 points and 1.6 ms for FAST-LIO2. This corresponds to a 25\% increase in the number of update points and a 68.8\% reduction in search time. In outdoor environments, GR-LIO processes about 2200 points with a search time of 4 ms, whereas FAST-LIO2 processes around 2000 points with a search time of 12 ms, resulting in a 10\% increase in utilized points and a 66.7\% reduction in search time. These results demonstrate that the proposed ground-based feature selection strategy significantly reduces the computational cost of feature association while increasing measurement utilization during state estimation.

This improvement arises from the incorporation of local ground plane priors in GR-LIO. For points classified as ground, observations are directly formed with the estimated local ground plane, thereby avoiding costly nearest-neighbor searches. In contrast, FAST-LIO2 relies on nearest-neighbor search for all points. Moreover, for points where FAST-LIO2 fails to establish valid correspondences (e.g., newly observed distant ground regions), our GR-LIO can still incorporate them into the state update process.

\subsubsection{Update Time}
We then evaluate the computational efficiency of the proposed method and the benchmark algorithms in both indoor and outdoor environments. Specifically, the average execution time per frame, including both the state update and the incremental map update, is measured for each sequence. The results are summarized in Table~\ref{tab:processing_time}.
GR-LIO (w/o PU) achieves the lowest average execution time among all evaluated methods in both indoor and outdoor environments. Even after enabling the planar motion update, GR-LIO remains more computationally efficient than FAST-LIO2 and the other benchmark algorithms. Notably, although GR-LIO incorporates more LiDAR points into the state update, it still achieves a lower overall execution time. This improvement is attributed to the proposed ground-aware observation model, which substantially reduces the computational cost of feature association, thereby offsetting the additional computation required to p rocess a larger number of observations. Furthermore, the planar motion update introduces only negligible computational overhead.

\begin{table}[t]
\centering
\caption{Average execution time per LiDAR frame, including state update and incremental map update (ms).}
\label{tab:processing_time}
\begin{tabular}{lcccc}
\toprule
Method & room 01 & hall 01 & gate 01 & street 03 \\
\midrule
LINS~\cite{c13} & 13.07 & 20.52 & 30.05 & 39.59 \\
LIO-SAM~\cite{c3}      & 8.65 & 23.31 & 36.01 & 55.71 \\
FAST-LIO2~\cite{c4}  & 3.46 & 8.07 & 20.76 & 19.07\\
GR-LIO (W/O PU) & \textbf{2.02} & \textbf{5.56} & \textbf{14.57} & \textbf{14.23} \\
GR-LIO       & \underline{2.24} & \underline{6.35} & \underline{15.40} & \underline{15.73} \\
\bottomrule
\end{tabular}
\begin{tablenotes}
\footnotesize
\item \textbf{Bold} and \underline{underlined} indicate the best and second-best results, respectively.
\end{tablenotes}
\end{table}

\subsection{Experiments on Self-collected Datasets}
\begin{figure}
\centering
    \subfloat[Room]{\includegraphics[width=0.95\linewidth]{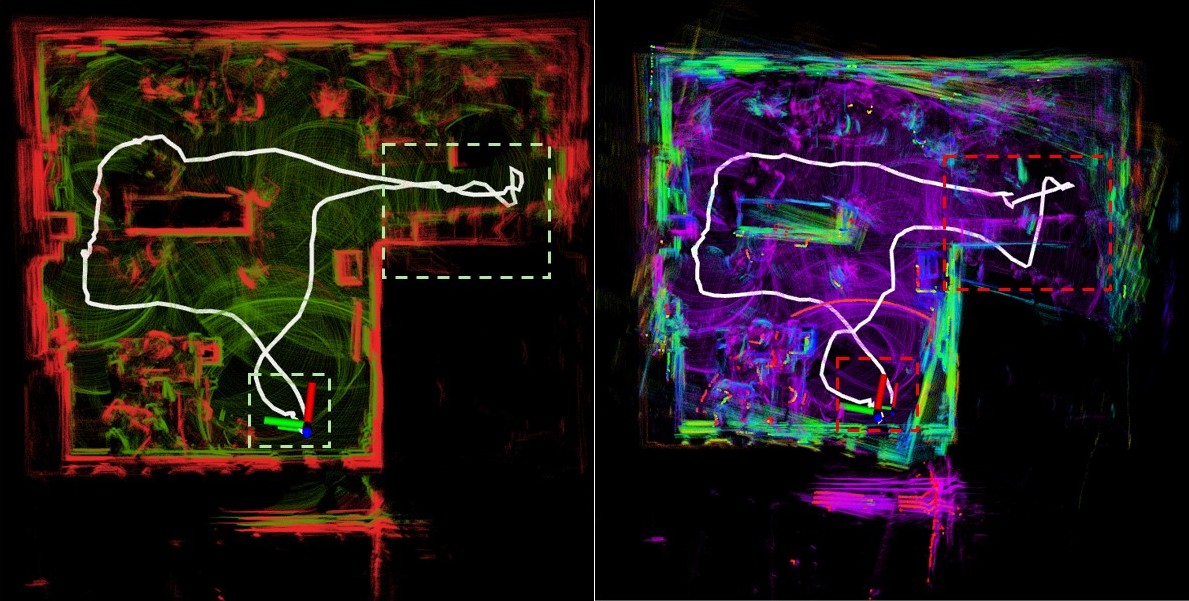}}
    \\
    \subfloat[Hall]{\includegraphics[width=0.95\linewidth]{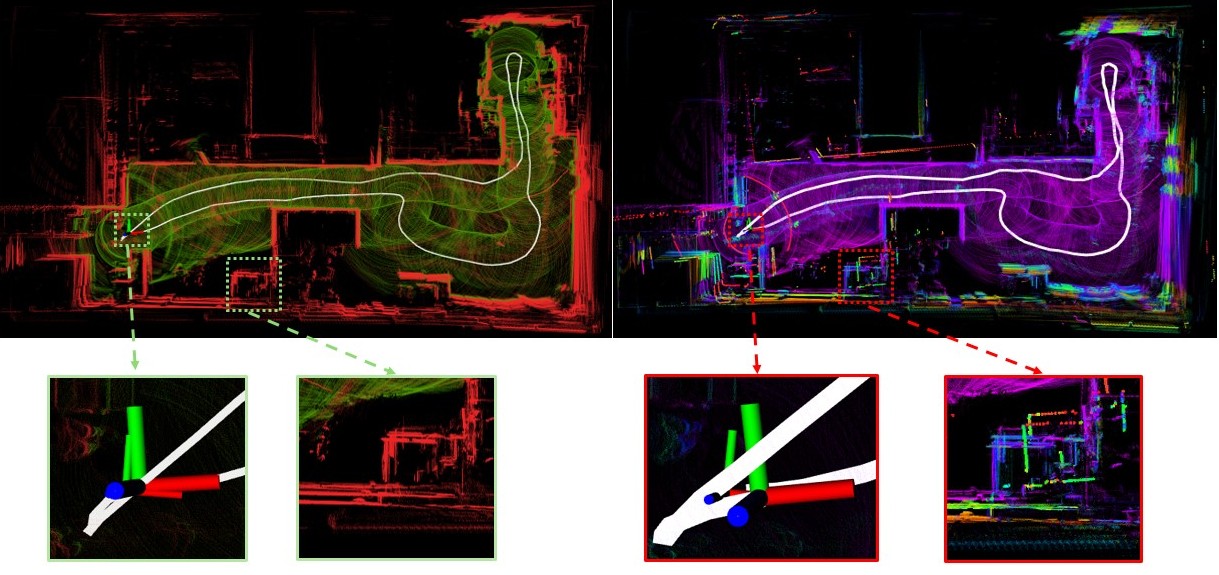}}
    \\ 
    \subfloat[Hall To Yard]{\includegraphics[width=0.95\linewidth]{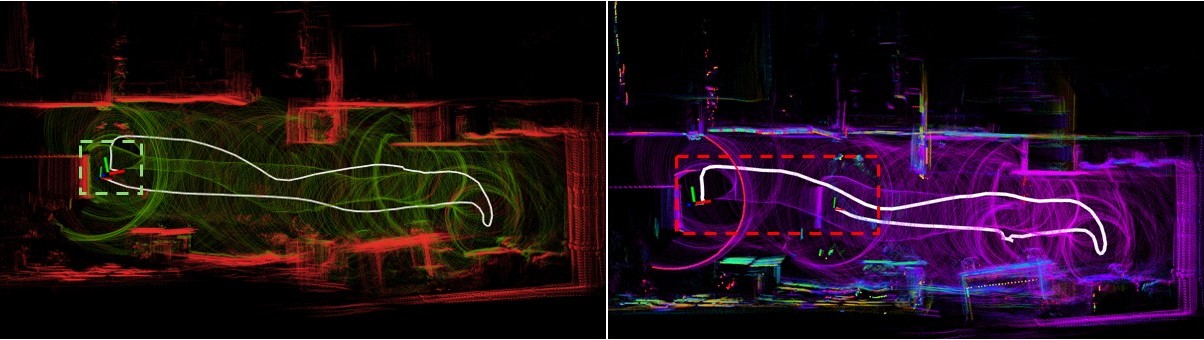}}
    \caption{Mapping results and estimated trajectories on the self-collected datasets. The white curves represent the estimated robot trajectories. The left and right columns correspond to GR-LIO and FAST-LIO2, respectively. In the results of our GR-LIO, segmented local ground points are highlighted in green, while non-ground points are shown in red. The coordinate frames indicate the estimated start and end poses of the robot.}
    \label{fig:selfdata}
\end{figure}

Fig.~\ref{fig:selfdata} presents the mapping results of GR-LIO and FAST-LIO2 on three representative self-collected sequences. To evaluate localization accuracy, the robot was driven along closed-loop trajectories with identical start and end locations. The accumulated error is assessed by measuring the discrepancy between the estimated final and the starting pose. 

The results show that our GR-LIO can efficiently segment local ground points and incorporate them into the state update process. Furthermore, GR-LIO exhibits smaller loop-closure drift than FAST-LIO2, as evidenced by the closer alignment between the estimated start and end poses. Specifically, the final loop-closure error is reduced from approximately 0.10~m to 0.02~m in the Room sequence and from 0.55~m to 0.17~m in the Hall sequence. In the Hall To Yard sequence, FAST-LIO2 exhibits a substantially larger loop-closure error, whereas GR-LIO still achieves a close alignment between the start and end poses. These results validate the effectiveness of exploiting local ground plane priors for improving localization accuracy and long-term consistency in ground robot navigation.

\section{CONCLUSIONS}
\label{sec:conclusion}
This paper presented GR-LIO, a ground-aware LiDAR-inertial odometry framework for ground robots. By parameterizing the local ground plane using the body-to-ground (B-G) height and the robot orientation, the proposed method incorporates local ground geometry into the state estimation process. To handle unknown B-G heights, a fast initialization method and an online calibration strategy are proposed to estimate and continuously refine this parameter during operation. Experimental results on multiple benchmark datasets demonstrate that the proposed method consistently improves both localization accuracy and computational efficiency. In particular, the proposed B-G parameterization enables efficient incorporation of ground observations into the state update, while the planar motion update effectively suppresses vertical drift and improves estimation accuracy. The proposed framework currently assumes locally planar ground surfaces. Future work will focus on extending the framework to more complex terrain by developing adaptive ground modeling and motion constraints for long-term  complex outdoor navigation.


\begin{thebibliography}{50}
\bibitem{c1} J. Zhang and S. Singh, “LOAM: Lidar odometry and mapping in real-time,” in Proc. Robotics: Science and Systems (RSS), Berkeley, CA, USA, Jul. 2014.
\bibitem{c2} J. Zhang and S. Singh, “Low-drift and real-time lidar odometry and mapping,” Auton. Robots, vol. 41, no. 2, pp. 401–416, 2017.
\bibitem{c3} T. Shan, B. Englot, D. Meyers, W. Wang, C. Ratti, and D. Rus, “LIO-SAM: Tightly-coupled lidar inertial odometry via smoothing and mapping,” in Proc. IEEE/RSJ Int. Conf. Intell. Robots Syst. (IROS), 2020, pp. 5135–5142.
\bibitem{c4} W. Xu, Y. Cai, D. He, J. Lin, and F. Zhang, “FAST-LIO2: Fast direct lidar-inertial odometry,” IEEE Trans. Robot., vol. 38, no. 4, pp. 2053–2073, 2022.
\bibitem{c5} T. Shan and B. Englot, “LeGO-LOAM: Lightweight and ground-optimized lidar odometry and mapping on variable terrain,” in Proc. IEEE/RSJ Int. Conf. Intell. Robots Syst. (IROS), 2018, pp. 4758–4765.
\bibitem{c6} A. Galeote-Luque, J.-R. Ruiz-Sarmiento, and J. Gonzalez-Jimenez, “GND-LO: Ground decoupled 3D LiDAR odometry based on planar patches,” IEEE Robot. Autom. Lett., vol. 8, no. 11, pp. 6923–6930, 2023.
\bibitem{c7} C. Forster, L. Carlone, F. Dellaert, and D. Scaramuzza, “On-manifold preintegration for real-time visual–inertial odometry,” IEEE Trans. Robot., vol. 33, no. 1, pp. 1–21, 2016.
\bibitem{c8} A. Segal, D. Haehnel, and S. Thrun, “Generalized-ICP,” in Proc. Robotics: Science and Systems (RSS), 2009.
\bibitem{c9} H. Ye, Y. Chen, and M. Liu, “Tightly coupled 3D LiDAR-inertial odometry and mapping,” in Proc. IEEE Int. Conf. Robot. Autom. (ICRA), 2019, pp. 3144–3150.
\bibitem{c10} P. Geneva, K. Eckenhoff, Y. Yang, and G. Huang, “LIPS: LiDAR-inertial 3D plane SLAM,” in Proc. IEEE/RSJ Int. Conf. Intell. Robots Syst. (IROS), 2018, pp. 123–130.
\bibitem{c11} T. Shan, B. Englot, C. Ratti, and D. Rus, “LVI-SAM: Tightly-coupled LiDAR–visual–inertial odometry via smoothing and mapping,” in Proc. IEEE Int. Conf. Robot. Autom. (ICRA), 2021, pp. 5692–5698.
\bibitem{c12} T. Wang, Y. Su, S. Shao, C. Yao, and Z. Wang, “GR-Fusion: Multi-sensor fusion SLAM for ground robots with high robustness and low drift,” in Proc. IEEE/RSJ Int. Conf. Intell. Robots Syst. (IROS), 2021, pp. 5440–5447.
\bibitem{c13} C. Qin, H. Ye, C. E. Pranata, J. Han, S. Zhang, and M. Liu, “LINS: A LiDAR-inertial state estimator for robust and efficient navigation,” in Proc. IEEE Int. Conf. Robot. Autom. (ICRA), 2020, pp. 8899–8906.
\bibitem{c14} W. Xu and F. Zhang, “FAST-LIO: A fast, robust LiDAR-inertial odometry package by tightly-coupled iterated Kalman filter,” IEEE Robot. Autom. Lett., vol. 6, no. 2, pp. 3317–3324, 2021.
\bibitem{c15} C. Bai, T. Xiao, Y. Chen, H. Wang, F. Zhang, and X. Gao, “Faster-LIO: Lightweight tightly coupled LiDAR-inertial odometry using parallel sparse incremental voxels,” IEEE Robot. Autom. Lett., vol. 7, no. 2, pp. 4861–4868, 2022.
\bibitem{c16} X. Zuo, P. Geneva, W. Lee, Y. Liu, and G. Huang, “LIC-Fusion: LiDAR-inertial-camera odometry,” in Proc. IEEE/RSJ Int. Conf. Intell. Robots Syst. (IROS), 2019, pp. 5848–5854.
\bibitem{c17} W. Lee, Y. Yang, and G. Huang, “Efficient multi-sensor aided inertial navigation with online calibration,” in Proc. IEEE Int. Conf. Robot. Autom. (ICRA), 2021, pp. 5706–5712.
\bibitem{c18} W. Lee, P. Geneva, C. Chen, and G. Huang, “MINS: Efficient and robust multisensor-aided inertial navigation system,” J. Field Robot., vol. 42, no. 7, pp. 3252–3284, 2025.
\bibitem{c19} V. Kubelka, L. Oswald, F. Pomerleau, F. Colas, T. Svoboda, and M. Reinstein, “Robust data fusion of multimodal sensory information for mobile robots,” J. Field Robot., vol. 32, no. 4, pp. 447–473, 2015.
\bibitem{c20} M. Himmelsbach, F. von Hundelshausen, and H.-J. Wuensche, “Fast segmentation of 3D point clouds for ground vehicles,” in Proc. IEEE Intell. Vehicles Symp. (IV), 2010, pp. 560–565.
\bibitem{c21} D. Zermas, I. Izzat, and N. Papanikolopoulos, “Fast segmentation of 3D point clouds: A paradigm on LiDAR data for autonomous vehicle applications,” in Proc. IEEE Int. Conf. Robot. Autom. (ICRA), 2017, pp. 5067–5073.
\bibitem{c22} Z. Yuan, X. Wang, J. Wu, J. Cheng, and X. Yang, “LiDAR-inertial odometry in dynamic driving scenarios using label consistency detection,” in Proc. IEEE/RSJ Int. Conf. Intell. Robots Syst. (IROS), 2025, pp. 1598–1605.
\bibitem{c23} Y. Su, T. Wang, S. Shao, C. Yao, and Z. Wang, “GR-LOAM: LiDAR-based sensor fusion SLAM for ground robots on complex terrain,” Robot. Auton. Syst., vol. 140, Art. no. 103759, 2021.
\bibitem{c24} K. J. Wu, C. X. Guo, G. Georgiou, and S. I. Roumeliotis, “VINS on wheels,” in Proc. IEEE Int. Conf. Robot. Autom. (ICRA), 2017, pp. 5155–5162.
\bibitem{c25} W. Li, G. Wang, Q. Zhang, and J. Liu, “LG-VIWO: Visual-Inertial-Wheel Odometry Leveraging the Depth-Aided Local Ground Constraints for Mobile Robots,” IEEE Trans. Instrum. Meas., vol. 74, pp. 1–11, 2025.
\bibitem{c251} Z. Zhang, W. Bai, L. Zhao, and P. Ladosz, “PL-VIWO: A lightweight and robust point-line monocular visual inertial wheel odometry,” in Proc. IEEE/RSJ Int. Conf. Intell. Robots Syst. (IROS), Hangzhou, China, 2025, pp. 15205–15212.
\bibitem{c252} Z. Zhang, L. Zhao, and P. Ladosz, “PL-VIWO2: A lightweight, fast and robust visual-inertial-wheel odometry using points and lines,” arXiv preprint arXiv:2509.21563, 2025.
\bibitem{c26} X. Wei, J. Lv, J. Sun, E. Dong, and S. Pu, “GCLO: Ground-constrained LiDAR odometry with low drift for GPS-denied indoor environments,” in Proc. IEEE Int. Conf. Robot. Autom. (ICRA), 2022, pp. 2229–2235.
\bibitem{c27} T. Guadagnino, B. Mersch, I. Vizzo, S. Gupta, M. V. R. Malladi, L. Lobefaro, G. Doisy, and C. Stachniss, “Kinematic-ICP: Enhancing LiDAR odometry with kinematic constraints for wheeled mobile robots moving on planar surfaces,” in Proc. IEEE Int. Conf. Robot. Autom. (ICRA), 2025, pp. 11090–11096.
\bibitem{c28} F. Pang, Y. Chen, Y. Li, Y. Zheng, and M. Luo, “Low-drift LiDAR-inertial odometry with reliable feature matching for unmanned ground vehicles,” IEEE Trans. Instrum. Meas., vol. 75, pp. 1–4, 2026.
\bibitem{c29} T. Kim, G. Pak, and E. Kim, “GRIL-Calib: Targetless ground robot IMU–LiDAR extrinsic calibration using ground plane motion constraints,” IEEE Robot. Autom. Lett., vol. 9, no. 6, pp. 5409–5416, 2024.
\bibitem{c291} M. A. Fischler and R. C. Bolles, ``Random sample consensus: A paradigm for model fitting with applications to image analysis and automated cartography,'' \emph{Communications of the ACM}, vol. 24, no. 6, pp. 381--395, 1981.
\bibitem{c30} J. Yin, A. Li, T. Li, W. Yu, and D. Zou, “M2DGR: A Multi-Sensor and Multi-Scenario SLAM Dataset for Ground Robots,” IEEE Robot. Autom. Lett., vol. 7, no. 2, pp. 2266–2273, 2022.
\bibitem{c31} D. Zhang, J. Zhang, Y. Sun, T. Li, H. Yin, H. Xie, and J. Yin, “Towards robust sensor-fusion ground SLAM: A comprehensive benchmark and a resilient framework,” in Proc. IEEE/RSJ Int. Conf. Intell. Robots Syst. (IROS), 2025, pp. 8894–8901.
\bibitem{c32}
M.~Grupp, ``EVO: Python package for the evaluation of odometry and SLAM,'' 2017. [Online]. Available: \url{https://github.com/MichaelGrupp/evo}.
\end{thebibliography}
\end{document}